\documentclass[runningheads]{llncs}

\usepackage{eccv}

\usepackage{eccvabbrv}

\usepackage{graphicx}
\usepackage{booktabs}
\usepackage{multirow}
\usepackage{bbding}
\usepackage{colortbl} 
\usepackage[dvipsnames]{xcolor}
\usepackage{array}
\usepackage{mathtools}
\usepackage{tabularx}
\usepackage[ruled,linesnumbered]{algorithm2e}
\usepackage{algpseudocode}

\usepackage[accsupp]{axessibility}  

\usepackage[pagebackref,breaklinks,colorlinks,citecolor=eccvblue]{hyperref}

\usepackage{orcidlink}

\begin{document}

\title{Omniview-Tuning: Boosting Viewpoint Invariance of Vision-Language Pre-training Models} 

\titlerunning{Omniview-Tuning}

\author{Shouwei Ruan\inst{1} \and
Yinpeng Dong\inst{2,4} \and
Hanqing Liu\inst{3} \and
Yao Huang\inst{1} \and
Hang Su\inst{2,5,6} \and
Xingxing Wei\inst{1,3}}

\authorrunning{R.~Showuei et al.}

\institute{
Institute of Artificial Intelligence, Beihang University, Beijing 100191, China \and
Dept. of Comp. Sci. \& Tech., Institute for AI, BNRist Center, THBI Lab,
Tsinghua-Bosch Joint ML Center, Tsinghua University, Beijing, 100084, China \and
Hangzhou Innovation Institute, Beihang University, Hangzhou 311228, China \and
RealAI \and Peng Cheng Laboratory \and Pazhou Laboratory (Huangpu), Guangzhou, China}

\maketitle

\begin{abstract}
Vision-Language Pre-training (VLP) models like CLIP have achieved remarkable success in computer vision and particularly demonstrated superior robustness to distribution shifts of 2D images. However, their robustness under 3D viewpoint variations is still limited, which can hinder the development for real-world applications. This paper successfully addresses this concern while keeping VLPs' original performance by breaking through two primary obstacles: 1) the scarcity of training data and 2) the suboptimal fine-tuning paradigms. To combat data scarcity, we build the Multi-View Caption (MVCap) dataset --- a comprehensive collection of over four million multi-view image-text pairs across more than 100K objects, providing more potential for VLP models to develop generalizable viewpoint-invariant representations. To address the limitations of existing paradigms in performance trade-offs and training efficiency, we design a novel fine-tuning framework named Omniview-Tuning (OVT). Specifically, OVT introduces a Cross-Viewpoint Alignment objective through a minimax-like optimization strategy, which effectively aligns representations of identical objects from diverse viewpoints without causing overfitting. Additionally, OVT fine-tunes VLP models in a parameter-efficient manner, leading to minimal computational cost. Extensive experiments on various VLP models with different architectures validate that OVT significantly improves the models' resilience to viewpoint shifts and keeps the original performance, establishing a pioneering standard for boosting the viewpoint invariance of VLP models.
  \keywords{Vision-Language Pre-training \and  Viewpoint Invariance}
\end{abstract}

\begin{figure}[htb]
  \centering
  \includegraphics[height=3.3cm]{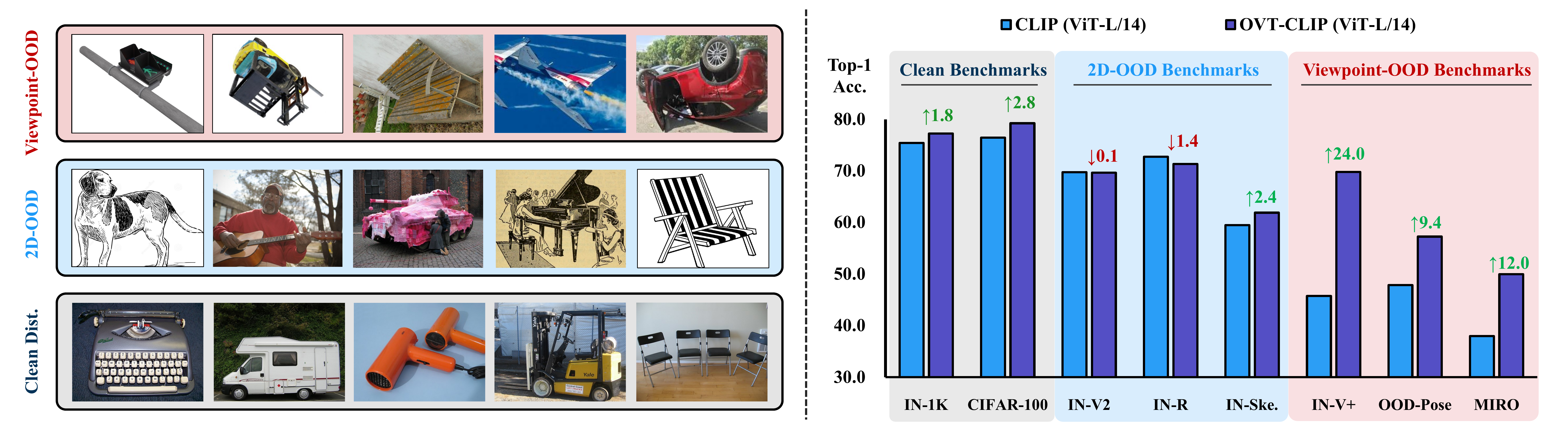}
  \caption{\textbf{The Challenge of Viewpoint Invariance in VLP.} We selected benchmarks representing clean distributions (ImageNet-1K~\cite{deng2009imagenet}, CIFAR-100~\cite{krizhevsky2009learning}), common 2D-OOD (ImageNet-V2~\cite{recht2019imagenet}, ImageNet-R(endition)~\cite{hendrycks2021many}, ImageNet-Sketch~\cite{wang2019learning}), and viewpoint-OOD (ImageNet-V(iewpoint)+~\cite{ruan2023towards}, OOD-CV(Pose)~\cite{zhao2022ood}, MIRO~\cite{cha2022miro}). We display samples from these data distributions (\emph{left}) and report the Top-1 accuracy of the original CLIP (ViT-L/14) and our improved OVT-CLIP (ViT-L/14) (\emph{right}). }

  \label{fig:cover}
  \vspace{-0.5cm}
\end{figure}




\section{Introduction}
\label{sec:intro}
Vision-Language Pre-training (VLP) models, such as CLIP~\cite{radford2021learning} and BLIP~\cite{li2022blip}, have shown great promise in learning transferable representations across various vision tasks. By aligning images and texts in a joint embedding space with a large corpus of paired image-text data,  
the VLP models exhibit exceptional representation and generalization capabilities that surpass traditional task-specific models. Owing to this, the VLP models serve as foundation models for numerous tasks, including visual recognition \cite{radford2021learning,jia2021scaling}, visual question answering~\cite{ liu2024visual,alayrac2022flamingo,zhu2023minigpt}, and text-to-image generation \cite{pmlr-v139-ramesh21a,saharia2022photorealistic}. Moreover, these models can effectively integrate real-world visual inputs with natural language instructions, leading to their increasing use in physical-world applications, like autonomous driving~\cite{zhou2023vision}, embodied robotics~\cite{wu2023embodied, li2023otter, huang2023voxposer}, \etc.


Besides their expressive power, VLP models have also shown excellent robustness under out-of-distribution (OOD) data~\cite{radford2021learning,pmlr-v162-fang22a,tu2023a}, including common corruptions~\cite{hendrycks2019benchmarking, calian2021defending, dong2023benchmarking}, stylistic changes~\cite{hendrycks2021many, wang2019learning}, and natural distribution shifts~\cite{recht2019imagenet, hendrycks2021many, hendrycks2021natural}.
However, 
a recent study~\cite{ruan2023improving} identifies that although VLP models excel at handling OOD data of 2D images, they suffer significant performance degradation under \textit{3D viewpoint changes}, revealing a notable shortcoming of the existing VLP models. 
As demonstrated in \cref{fig:cover}, when dealing with the recently introduced benchmarks concerned with 3D viewpoint shifts~\cite{dong2022viewfool,ruan2023towards,zhao2022ood}, CLIP's performance is obviously lower than that on 2D-OOD benchmarks. This large gap likely stems from limited coverage of diverse viewpoints in the training datasets~\cite{thomee2016yfcc100m, schuhmann2022laion, gadre2023datacomp}, which is crucial for learning viewpoint-invariant representations. As VLP models are increasingly deployed in real-world environments where viewpoint shifts often occur, enhancing their resilience to such changes is urgent and essential.

\begin{figure}[t]
  \centering
  \includegraphics[height=10.3cm]{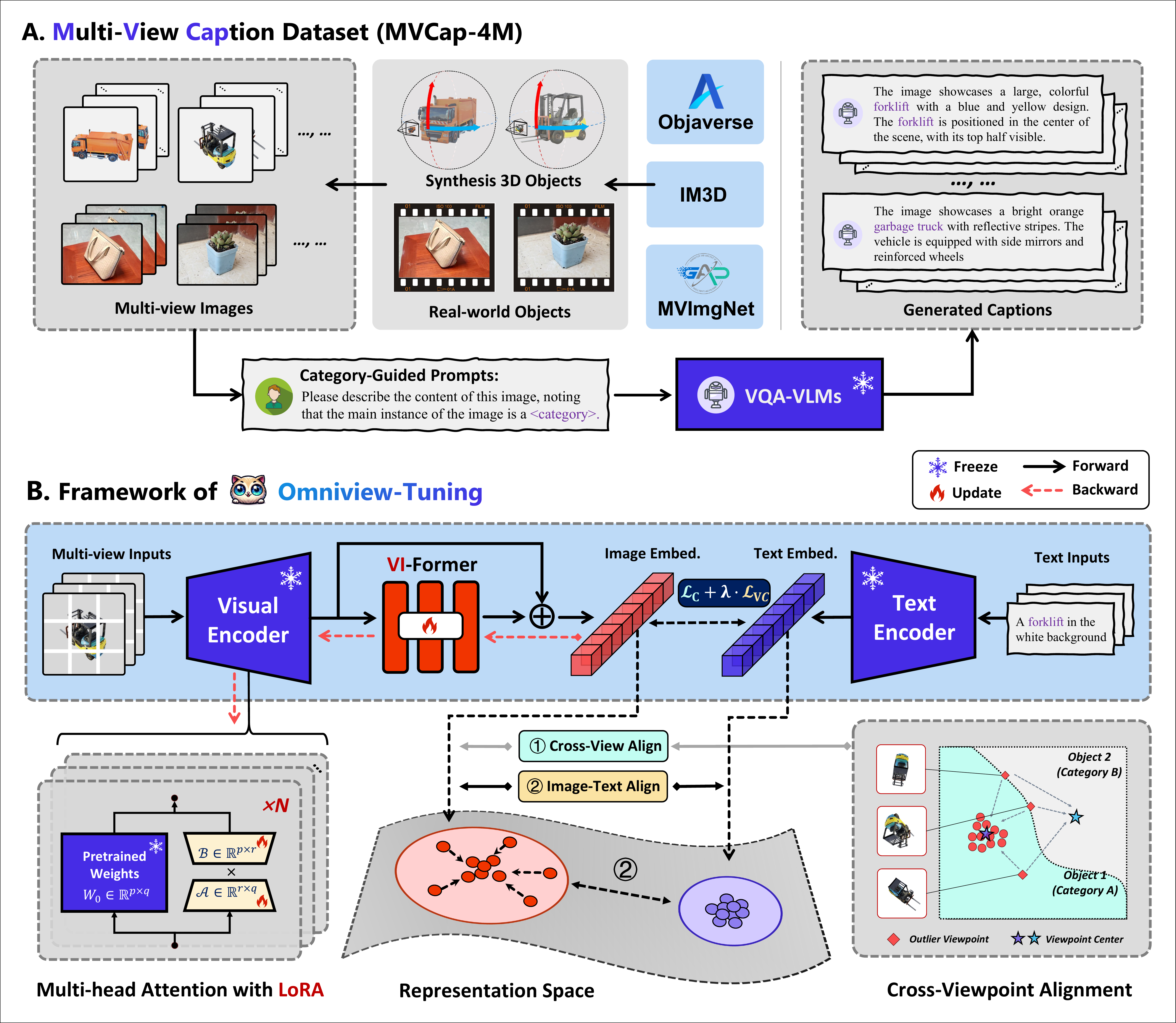}
  \caption{\textbf{Method Overview.} \textbf{(A)} We create the first multi-view image caption dataset by collecting multi-view samples from existing 3D object and video datasets, and generating category-guided descriptions using VLLMs. \textbf{(B)} The proposed Omniview-Tuning takes multi-view image caption data as input, employs the cross-view alignment objective to encourage the model to learn viewpoint-invariant representations, and achieves efficient fine-tuning by updating VIformer and LoRA parameters.
  }
  \label{fig:framework}
  \vspace{-0.5cm}
\end{figure}

To address this problem, this paper sets out to \emph{enhance the viewpoint invariance of VLP models while preserving the original performance as much as possible}. However, achieving this goal meets the following challenges: (1) \emph{Data scarcity}: acquiring VLP training data that covers a wide range of viewpoint variations is particularly challenging compared to conventional image-text pair data. Although some datasets introduced for task-specific models do include viewpoint variations~\cite{barbu2019objectnet,madan2020and,hendrycks2021natural,zhao2022ood}, they often lack the textual descriptions vital for VLP. Even the largest available multi-view datasets~\cite{ruan2023towards,collins2022abo,yu2023mvimgnet} fall short in terms of scale, category coverage, and viewpoint diversity, thereby limiting the potential for VLP models to develop generalizable viewpoint-invariant representations. (2) \emph{Inappropriate paradigms}: traditional approaches, which often regard viewpoint changes as adversarial attacks and employ adversarial training paradigms for enhancing invariance~\cite{alcorn2019strike, ruan2023towards, ruan2023improving}, are not entirely suitable for VLP models. Such frameworks typically entail a trade-off between robustness and accuracy—a balance that requires more careful consideration for foundation VLP models, where our aim is not solely to improve viewpoint invariance but, more importantly, to bridge the gap between it and the original performance. Furthermore, these approaches necessitate extra 3D reconstruction and neural rendering to capture adversarial viewpoints, leading to prohibitive computational costs for large-scale VLP models. For instance, tuning ResNet-50 with VIAT~\cite{ruan2023towards} under a dataset of just 1K objects demands around 400 GPU hours. Therefore, it is important to make training more efficient and less resource-intensive.

Based on the above discussions, this paper conducts a pioneering exploration of the viewpoint invariance of VLP models. Specifically, we address the aforementioned challenges by making the following contributions:

\textbf{\emph{Million-scale multi-view image-text training set.}} We introduce a large-scale \textbf{M}ulti-\textbf{V}iew \textbf{Cap}tion (\textbf{MVCap}) dataset tailored for viewpoint invariance of VLP models, comprising over 4.6 million multi-view image-text pairs across more than 100K objects. To assemble a diverse collection of multi-view image-text pairs, we amalgamate various 3D assets with real-world multi-view data. This process involves an extensive selection and rendering of multi-view images from existing datasets. 
We then utilize a Vision Large Language Model (VLLM) for automated caption generation to obtain semantically rich textual descriptions without extensive manual efforts.
To ensure category consistency across varying viewpoints in the generated captions, we implement a category-guided prompting strategy, which maintains accuracy in textual descriptions for different viewpoints of the same object or scene (details in~\cref{sec:dataset}). 

\textbf{\emph{Effective framework for enhancing VLP's viewpoint invariance.}} We propose \textbf{Omniview-Tuning (OVT)}, a novel framework designed to enhance the viewpoint invariance of prevalent VLP models. As illustrated in~\cref{fig:framework}, OVT employs multi-view image-text pairs for training additional learnable components. To amplify the model's proficiency in learning viewpoint-invariant representations, we introduce a \textbf{Cross-viewpoint Alignment} objective, ensuring that representations of the same object from different viewpoints are close and unified in the high-dimensional feature space. To prevent performance trade-offs due to the concept drift from aggressive viewpoint alignment, we innovatively construct the optimization paradigm of OVT in a \textbf{minimax-like form}. The optimization process includes identifying extreme outlier viewpoints during the maximization step, while optimizing the model's invariant representation for these outlier samples in the minimization step. This strategy enables the model to focus more on the worst-case viewpoint samples, thereby maximally preserving the original embedding distribution and avoiding performance degradation while saving computational costs. Moreover, OVT is designed in a \textbf{Parameter-Efficient} Fine-Tuning manner to improve efficiency, and creatively incorporates two trainable parameter modules: an embedding transformation module named VIFormer and the Low-Rank Adaptation (LoRA~\cite{hu2021lora}) weights, to acquire additional viewpoint invariance capabilities efficiently.

\textbf{\emph{Extensive experiments across various VLP architectures and tasks.}} We conduct extensive experiments to show the efficacy of the OVT framework in improving the viewpoint invariance for VLP models while maintaining performance on clean data and 2D-OOD samples. For example, by fine-tuning CLIP with OVT on different architectures (ViT-B/32, ViT-B/16, and ViT-L/14), the Top-1 accuracy on viewpoint-OOD benchmarks increased by an average of \textbf{9.6\%}, \textbf{10.2\%}, and \textbf{8.9\%}, respectively, with only a minimal sacrifice on 2D-OOD benchmarks by an average of 2.6\%, 1.4\%, and 0.2\%. Furthermore, serving as the visual encoder in VLLMs (\eg, LLaVa \cite{liu2024visual}), OVT-CLIP also effectively improves viewpoint invariance in image captioning and visual question answering tasks.

\section{Related Work}

\subsection{Viewpoint Invariance and Robustness}
Viewpoint invariance is a key property of human vision \cite{biederman1987recognition} but is usually lacking in computer vision models \cite{alcorn2019strike,dong2022viewfool}.
Addressing viewpoint invariance and robustness involves strategies like data augmentation and adversarial learning. Early efforts aim to enhance viewpoint robustness by incorporating datasets enriched with viewpoint variations~\cite{barbu2019objectnet,madan2020and,hendrycks2021natural,zhao2022ood}. For example, Madan~\etal encourage models to learn viewpoint-robust representations by incorporating object-pose combinations~\cite{madan2020and}. However, these methods often falter under malicious viewpoint perturbations due to their inability to capture the worst-case viewpoint samples. Recently, achieving viewpoint invariance within the adversarial training paradigm has shown promise~\cite{alcorn2019strike, hamdi2020towards, dong2022viewfool, ruan2023towards}. By treating viewpoint variations as an adversarial attack, Alcorn~\etal employ a differentiable renderer to train models against adversarial viewpoints optimized from a limited 3D objects set~\cite{alcorn2019strike}. Recent studies, such as Viewfool~\cite{dong2022viewfool} and VIAT~\cite{ruan2023towards, ruan2023improving}, have introduced neural radiance field (NeRF)~\cite{mildenhall2021nerf, muller2022instant}, enabling the characterization of adversarial viewpoint distributions from 2D multi-view inputs. Notably, VIAT adopts adversarial distribution training, significantly improves viewpoint invariance, and successfully generalizes performance to unseen objects. Distinct from previous studies, our work pioneers the improvement of viewpoint invariance representation within large-scale VLP models, which is facilitated through suitable training data and refined fine-tuning methodologies.

\subsection{Vision-Language Pre-training}
In the realm of VLP, significant strides have been made in understanding and bridging the semantic gap between visual and textual information. Despite the variety of existing VLP paradigms, such as single-stream encoder (\eg, VisualBERT~\cite{li2019visualbert} and UNITER~\cite{chen2020uniter}, \etc) or dual-stream encoder equipped with diverse training objectives, the dual-stream contrastive learning architecture exemplified by ALIGN~\cite{li2021align} and OpenAI’s CLIP~\cite{radford2021learning} dominates the field. CLIP, in particular, has gained widespread attention for its ability to perform zero-shot classification tasks by adopting a vast corpus of internet-collected image-text pairs, demonstrating the power of large-scale contrastive pre-training. Thus, Our investigation primarily focuses on these VLP architectures. Building upon these foundational works, subsequent iterations like open-CLIP~\cite{ilharco_gabriel_2021_5143773}, EVA-CLIP~\cite{sun2023eva,sun2024eva}, and MetaCLIP~\cite{xu2023demystifying} have introduced nuanced enhancements. These refinements, ranging from the incorporation of more expansive high-quality image-text datasets and improved training methodologies, have collectively contributed to performance uplifts. BLIP~\cite{li2022blip}, meanwhile, introduces a bootstrapping mechanism by the proposed captioner and ﬁlter module that achieve significant performance improvements on various downstream tasks.

\section{Multi-view Caption Dataset}\label{sec:dataset}

We recognize that one of the key challenges in achieving viewpoint invariance for VLP is the scarcity of training data that offer comprehensive viewpoint sampling. As summarized in~\cref{table:dataset}, existing large-scale multi-view datasets~\cite{ho2019catastrophic,reizenstein2021common,collins2022abo,yu2023mvimgnet,ruan2023towards} typically lack in either sample diversity, category breadth, or textual descriptions, limiting their effectiveness for supporting VLP models to achieve viewpoint invariance. To address these limitations, we introduce the MVCap dataset. The subsequent sections detail its construction.

\begin{table*}[t]
\caption{\textbf{Comparison of current large-scale multi-view datasets.} \emph{"Spherical"} indicates whether the viewpoints cover spherical space, \emph{"Diversity"} assesses the diversity of viewpoints, and \emph{"Caption"} indicates whether textual descriptions are provided.}
\vspace{-0.2cm}
\tiny
\setlength\tabcolsep{2.7pt}
\renewcommand\arraystretch{1.2}
\centering
\begin{tabular}{l|c|c|c|c|c|c|c|c|c}
\hline
Dataset      & Year & \#Obj. & \#Cat. & \#Avg.~View. & \#Sample & Image Domain   & Spherical & Diversity & Caption \\ \hline\hline
OOWL~\cite{ho2019catastrophic}         & 2019 & 500      & 25          & 240             & 120K      & Real           & \XSolidBrush              & $\bigstar$$\bigstar$                  & \XSolidBrush        \\
CO3D~\cite{reizenstein2021common}         & 2021 & 18.6k    & 50          & $\sim$80              & 1.5M      & Real           & \XSolidBrush              & $\bigstar$                   & \XSolidBrush        \\
ABO~\cite{collins2022abo}          & 2022 & 7.9k     & 63          & 30              & 238K      & Synthetic      & \XSolidBrush             & $\bigstar$                   & \XSolidBrush        \\
IM3D~\cite{ruan2023towards}         & 2023 & 1.0k     & 100         & 100             & 100K      & Synthetic      & \Checkmark              & $\bigstar$$\bigstar$$\bigstar$                 & \XSolidBrush        \\
MVImgNet~\cite{yu2023mvimgnet}     & 2023 & 219.0k   & 238         & $\sim$30              & 6.5M      & Real           & \XSolidBrush             & $\bigstar$$\bigstar$                  & \XSolidBrush        \\
\rowcolor{gray!25} MVCap (Ours) & 2024 & 94.6k    & 1600        & 100/$\sim$30          & 4.6M      & Synthetic+Real & \Checkmark              & $\bigstar$$\bigstar$$\bigstar$                 & \Checkmark        \\ \hline
\end{tabular}
\vspace{-0.3cm}
\label{table:dataset}
\end{table*}

\begin{figure}[t]
  \centering
  \includegraphics[height=2.3cm]{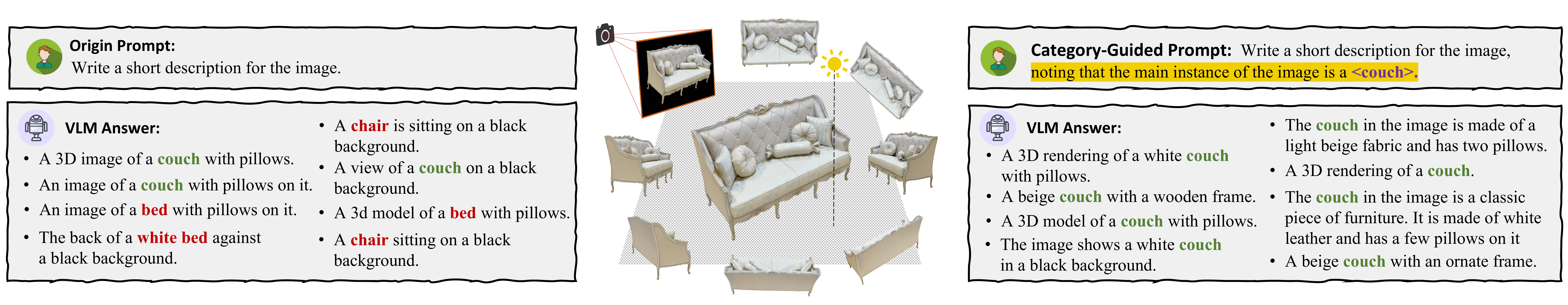}
  \vspace{-0.2cm}
  \caption{Generated multi-view captions with common and category-guided prompts.
  }
  \label{fig:dataset}
  \vspace{-0.3cm}
\end{figure}

\subsection{Multi-View Image Collection} 
We commence by gathering a multi-view image collection $\mathcal{D} =\{I_{ij}\mid i\!=\!1,2,...,N;j\!=\!1,2,...,M_i\}$, where $N$ and $M_i$ represent the counts of objects and their viewpoints, respectively. To cover various categories from virtual to real-world scenes, we integrate samples from Objaverse~\cite{deitke2023objaverse}, IM3D~\cite{ruan2023towards}, and MVImgNet~\cite{yu2023mvimgnet}. Since the original 3D dataset includes a fair share of noisy and semantically indistinct objects, we leverage semantic embeddings provided by OpenShape~\cite{liu2024openshape} to conduct cosine similarity sorting based on the embeddings of customized labels. Finally, we filter 24,495 virtual 3D objects endowed with distinct semantic clarity and cover over 1,600 categories. For each chosen 3D object, we employ Blender to render 100 random viewpoint images from the upper hemisphere, ensuring a comprehensive and varied viewpoint representation in our collected samples. We also incorporate objects from MVImgNet with over 30 valid viewpoints (video frames), thereby acquiring a substantial number of real-world multi-view samples to enrich the dataset's content and quality further.

\subsection{Category-Guided Caption Generation} 
The granularity and precision of textual descriptions are pivotal in VLP training, as they influence the model's generalization capabilities and the variety of visual concepts learned. Relying solely on simple prompt engineering, such as "\emph{a photo of [category]}," may introduce biases and limit the model's generalizability, whereas manual annotation is costly. To circumvent this, we utilize InstructBLIP-flant5xl~\cite{instructblip}, a leading VLLM, to create multi-view captions automatically. However, such VLLMs also grapple with viewpoint invariance, where the model's responses to different viewpoints can often be category-inconsistent, as depicted in \cref{fig:dataset}. This situation presents a "chicken or egg" dilemma: we hope to use a viewpoint-invariant model to supply data for viewpoint invariance training. We address this by the design of category-guided prompting. Specifically, we use prompts containing ground-truth category information to eliminate the hallucination of large VLLMs in response to viewpoint-shifted inputs, thereby generating category-consistent multi-view captions. Formally, the forward process for generating captions can be represented as follows:
\begin{equation}
\begin{array}{c}
    T_{ij}= \mathcal{G}[I_{ij}, \textrm{Prompt}(c_i)];\vspace{0.2cm} \\ 
    \textrm{Prompt}(c_i)=\textrm{"\emph{Write a short description for the image}},\\ 
    \textrm{\emph{noting that the main instance of the image is a}} <c_i> .\textrm{"},
    \label{eq: 1}
\end{array}
\end{equation}
where $c_i \in C$ denotes the category label for the $i$-th object, and $\mathcal{G}$ denotes the forward process of InstructBLIP. This yields the multi-view image-text pairs $\tilde{\mathcal{D}} =\{\left \langle I_{ij}, T_{ij} \right \rangle \mid i=1,2,...,N;j=1,2,...,M_i\}$, which can be utilized for the viewpoint invariance fine-tuning of VLP models.

\section{Omniview-Tuning}
Similar to traditional task-specific visual models~\cite{dong2022viewfool}, VLP models are equally vulnerable to viewpoint variations, necessitating research into their viewpoint invariance enhancement. Next, we will first review the paradigm of VLP in~\cref{sec:preliminaries}. Building on this, we present the problem formulation of OVT in~\cref{sec:Problem Formulation} and detail the specific training techniques in~\cref{sec:Min-Max Optimization Strategy} and~\cref{sec:Parameter-Efficient Modules}.

\subsection{Preliminaries: Contrastive Vision-Language Pre-training}\label{sec:preliminaries}
Despite the variety of existing VLP paradigms, such as single-stream or dual-stream architectures equipped with diverse training objectives, the dual-stream contrastive learning architecture exemplified by CLIP \etal~\cite{radford2021learning,li2021align} dominates the field. Thus, Our investigation primarily focuses on this VLP architecture. 

Without the loss of generality, these VLP models are composed of a visual encoder $E_{\mathbf{W_v}}:I\rightarrow z^I\in\mathbb{R}^d$ and a text encoder $E_{\mathbf{W_t}}:T\rightarrow z^T\in\mathbb{R}^d$, which maps visual and textual inputs to a unified high-dimensional feature space ${\mathbb{R}}^d$, respectively, where $\mathbf{W_v}$ and $\mathbf{W_t}$ are weight matrices of two encoders. Given a large corpus of image-text pairs $\{\left \langle I_{i}, T_{i} \right \rangle\}_{i=1}^N$, VLP models typically employ an image-text contrastive (ITC) loss as the training objective:
\begin{equation}
\begin{array}{c} 
\mathcal{L}_{ITC} =  \frac{1}{2} (\mathcal{L}_{I\rightarrow T}+\mathcal{L}_{T\rightarrow I}),\\ 
\end{array}
\label{eq:itc loss}
\end{equation}
\noindent which is composed of an image-to-text and a text-to-image terms formulated as:
\begin{equation}
\begin{array}{c} 
\mathcal{L}_{I\rightarrow T}=-\frac{1}{N}\sum_{i=1}^{N}\log\frac{\exp(d(z^I_{i},z^T_{i})/\tau )}{\sum_{k=1}^{N}\exp(d(z^I_{i},z^T_{k})/\tau )},\vspace{0.2cm} \\ 
\mathcal{L}_{T\rightarrow I}=-\frac{1}{N}\sum_{i=1}^{N}\log\frac{\exp(d(z^T_{i},z^I_{i})/\tau )}{\sum_{k=1}^{N}\exp(d(z^T_{i},z^I_{k})/\tau )},
\end{array}
\label{eq:itc loss parts}
\end{equation}
\noindent where $\tau$ represents a learnable temperature parameter, $z^I$ and $z^T$ denote the image and text embeddings, respectively. The $\mathcal{L}_{ITC}$ maximizes the similarity between matched image-text pairs while minimizing the similarity for mismatched pairs, thus enabling the alignment of visual and textual information to the same feature space, bringing the embeddings of matched pairs closer. Following~\cite{radford2021learning,li2021align,li2022blip}, the proposed Omniview-Tuning implements $\mathcal{L}_{ITC}$ for aligning multi-view images with text modalities, which is explained in the next section.

\subsection{Problem Formulation} \label{sec:Problem Formulation}
\noindent \textbf{Viewpoint Invariance of Vision-Language Pre-training.} In computer vision scenario, viewpoint invariance implies that model $f(\cdot)$ can provide consistent predictions or representations given any different views of the identical object or scene~\cite{ruan2023towards}. Formally,  given a collection of multi-view images $\mathcal{D} =\{I_{ij}\mid i\!=\!1,2,...,N;j\!=\!1,2,...,M_i\}$, viewpoint invariance is required:
\begin{equation}
    f(I_{ij})=f(I_{ij' }), \quad \forall i, j, j' ~\textrm{with}~j\neq j',
\end{equation}
\noindent where $i$ is the index of the object/scene, $j$ and $j'$ are indexes of two viewpoint samples. However, in the context of dual-stream VLP models, this concept requires a more refined interpretation. For VLP models, viewpoint invariance necessitates that the visual representations (\ie, the embeddings inferred from the visual encoder) from different viewpoints be sufficiently close in the feature space. Assuming $I_{ij}$ and $I_{ij'}$ are images from different viewpoints of the same object, this requirement can be formulated as follows:
\begin{equation}
\small
    d \Big[E_{\mathbf{W_v}}(I_{ij}), E_{\mathbf{W_v}}(I_{ij'}) \Big]\le \epsilon , 
\end{equation}
where $d(\cdot)$ denotes a distance metric in the representation space, such as cosine distance, $\epsilon$ represents the maximum variance allowed. 

\textbf{Optimization Objectives of Omniview-Tuning.}  Although images from different viewpoints often correspond to slightly varying textual descriptions, influenced by context, grammatical structure, and linguistic ambiguity, this variation could be significantly amplified in the high-dimensional representation space~\cite{rong2014word2vec,chuang2022diffcse}. Therefore, relying solely on image-text alignment may not suffice to adequately align embeddings from different viewpoints. Starting from the definition of viewpoint invariance, we introduce a cross-viewpoint alignment objective within the $\mathcal{L}_{ITC}$ to directly encourage the model to learn invariant representations between different viewpoints, rather than relying on the indirect alignment through textual descriptions. This can be seen as a regularization that forces the model to obtain viewpoint invariance, even when such invariance is not explicitly articulated in the textual descriptions. With this consideration, given a multi-view training set $\mathcal{D}$, the optimization problem is defined as follows:

\begin{equation}
    \min_{\mathbf{W_v},\mathbf{W_t}} \big [\mathcal{L}_{ITC} + \lambda\cdot \underbracket[1pt][3pt]{\textstyle\sum_i \textstyle\sum_{j\neq j'} d(z^I_{ij},z^I_{ij'})}_{\mathcal{L}_{VC} } \big] , 
\label{eq:ovt problem}
\end{equation}
where the first term represents the image-text alignment used in the pre-training process, while the second term signifies the cross-viewpoint alignment goal mentioned above, referred to as Viewpoint Consistency loss ($\mathcal{L}_{VC}$), which aims to minimize the cosine distance between embeddings from different viewpoints. $\lambda$ is a hyperparameter that balances the importance of two loss terms. 

\subsection{Optimization Strategy} \label{sec:Min-Max Optimization Strategy}
In summary, the naive way to achieve viewpoint invariance is to calculate the loss terms in~\cref{eq:ovt problem} based on the forward process of encoders, then update the encoders' weight using gradient descent. However, it has a relatively high time complexity to solve \cref{eq:ovt problem} because current $\mathcal{L}_{VC}$ requires iterating over every possible combination of viewpoints. Therefore, we endeavor to provide a more effective implementation  for the original optimization problem. Drawing from the advantages of adversarial training \cite{madry2017towards,ruan2023towards}, we frame the optimization of the $\mathcal{L}_{VC}$ in a minimax format, rewriting the original problem~\cref{eq:ovt problem} as: 
\begin{equation}
\small
\begin{array}{c}
\min_{\mathbf{W_v},\mathbf{W_t}}\Biggr[~\mathcal{L}_{ITC}+\lambda\cdot \underbracket[1pt][3pt]{\max_{\mathcal{O}=\{O_i\}_{i=1}^N, \left |  O_i \right |=K }  {\textstyle \sum_{i=1}^{N} {\textstyle \sum_{j\in\mathcal{O} }^{} l(z^I_{ij},z^I_{C_i})} }}_{\mathcal{L}_{VC} } ~\Biggr] , \vspace{0.2cm} \\
\mathrm{where}~~l(z^I_{ij},z^I_{C_i})= \max\big[d(z^I_{ij},z^I_{C_i})+m, 0\big],
\end{array}
\label{eq:problem}
\end{equation}
where $\mathcal{O}\!=\!\{O_i\}_{i=1}^N$ is the outlier viewpoints set, $z^I_{C_i}$ are anchor viewpoint embeddings of each object, and $l(\cdot)$ is the cosine distance with a margin $m$. During the optimization, The maximization step first identifies the collection of top-$K$ outlier viewpoints $\mathcal{O}$, which are the viewpoint samples with the highest degree of representational deviation. Then, the minimization step encourages the outlier viewpoint embeddings to converge towards corresponding anchor viewpoint embeddings $z^I_{C_i}$. We obtain $z^I_{C_i}$ by calculating the \textbf{nearest-neighbor weighted} embedding centroid of each object:
\begin{equation}
\begin{array}{c}
z^I_{C_i}= {\textstyle \sum_{j=1}^{M_i}\tilde{\omega}_{ij} \cdot z^I_{ij}},\vspace{0.2cm} \\ \mathrm{where}~~\tilde{\omega}_{ij}=\omega_{ij}/\textstyle \sum_j \omega_{ij},~~\omega_{ij}=1/ {\textstyle \sum_{z^I_{ih}\in \mathcal{Q}_{ij}}^{} d(z^I_{ij},z^I_{ih}) } 
\end{array}
\end{equation}
\noindent where $\mathcal{Q}_{ij}=\{z^I_{ih}\}_{h=1}^5$ is the top-5 nearest neighbours of each viewpoint embedding $z^I_{ij}$. As for the outlier viewpoints set, we define them as the viewpoints with the top-$K$ farthest cosine distances from $z^I_{C_i}$. 

The adoption of this strategy offers dual advantages: \emph{Firstly}, it allows the model to focus solely on extreme outlier viewpoints, preventing concept drift and potential overfitting to the fine-tuning dataset that results from excessive alignment. \emph{Secondly}, this approach reduces computational overhead and significantly enhances optimization efficiency.

\subsection{Parameter-Efficient Modules} \label{sec:Parameter-Efficient Modules}

To mitigate the impact of full parameters update on the original performance and enhance training efficiency, we achieve viewpoint invariance by efficiently fine-tuning the parameters of the visual encoder while keeping the text encoder frozen. Inspired by LoRA~\cite{hu2021lora}, we perform low-rank decomposition on the weights of the visual encoder $\mathbf{W_v}\in \mathbb{R}^{n\times m}$ to substitute full-parameter update: 
\begin{equation}
\small
    \tilde{\mathbf{W_v}}=\mathbf{W_v}+\Delta W=\mathbf{W_v}+\mathbf{BA},~~\textrm{where}~\mathbf{B}\in\mathbb{R}^{m\times r},\mathbf{A}\in\mathbb{R}^{r\times n},r\ll \min(n,m),
\end{equation}
where $\mathbf{A}$ and $\mathbf{B}$ are two learnable low-rank parameter matrices, which we apply to the self-attention layers of the visual encoder and update them during fine-tuning while freezing the original pre-trained weights. This enables us to enhance the model's viewpoint invariance representation capability with minor parameter changes while maximizing the preservation of the original performance. 
Drawing inspiration from the success of CLIP-Adapter~\cite{gao2023clip}, which improves CLIP's performance in few-shot scenarios by introducing linear layers after the encoder, we propose a similar module called VIformer:$f_{\boldsymbol{\theta}}:z^I\in\mathbb{R}^d\rightarrow s^I\in\mathbb{R}^d$, where $\boldsymbol{\theta}$ is the weight. Unlike CLIP-Adapter, VIformer transforms the original embeddings $z^I$ by introducing self-attention layers in a learnable manner to extract and retain specific viewpoint-invariant key components $s^I$. Combining LoRA and VIformer modules, the forward process of image encoding can be represented as follows: 
\begin{equation}
\begin{split} 
  \tilde{z}^I &=\alpha \cdot f_{\boldsymbol{\theta}} (z^I)+(1-\alpha )\cdot z^I \\
   \vspace{0.2cm}
   &=\alpha \cdot f_{\boldsymbol{\theta}} (\mathbf{W_v}\cdot I+\mathbf{BA}\cdot I)+(1-\alpha )\cdot (\mathbf{W_v}\cdot I+\mathbf{BA}\cdot I),
\end{split}
\label{eq:forward}
\end{equation}
where the constant value $\alpha$ denotes the residual ratio to balance achieving original performance and viewpoint invariance performance. Therefore, for~\cref{eq:ovt problem}, we now only need to update $\mathbf{A}$, $\mathbf{B}$, and $\boldsymbol{\theta}$, rather than the entire weights of VLP.

In practice, we combine MVCap and ImageNet-1K training set to fine-tune the network. Before each epoch, we first calculate the anchor viewpoints for all objects in the dataset and perform the maximization process to compute the set of outlier viewpoints collection. Then, we calculate the $\mathcal{L}_{ITC}$ and $\mathcal{L}_{VC}$ in each batch and update the $\mathbf{A}$, $\mathbf{B}$, and $\boldsymbol{\theta}$ through gradient descent.

\begin{table*}[t]
\caption{\textbf{Configurations of OVT and zero-shot Top-1 accuracy (\%) on ImageNet-1K with ImageNet-V+.} The number in parentheses shows the performance change relative to the pre-trained weights. Through OVT training, each model maintains the performance on ImageNet-1K (IN-1K) while significantly improving the performance on ImageNet-V+ (IN-V+.), narrowing the performance gap.}
\vspace{-0.1cm}
\setlength\tabcolsep{1.8pt}
\renewcommand\arraystretch{1.3}
\centering
\resizebox{\textwidth}{!}{\begin{tabular}{l|lc|cc|ccc|cc}
\hline
\multicolumn{1}{c|}{Model} & \multicolumn{1}{c}{Pretrain Weight} & \multicolumn{1}{c|}{Pretrain Data} & \begin{tabular}[c]{@{}c@{}} Total\\\#Param.\end{tabular} & \begin{tabular}[c]{@{}c@{}}Trainable \\ \#Param.\end{tabular} & \begin{tabular}[c]{@{}c@{}} Image\\Size\end{tabular} & \begin{tabular}[c]{@{}c@{}} Batch\\Size\end{tabular} & \#Iter. & IN-1K & IN-View+ \\ \hline \hline
\textbf{\textcolor{BlueViolet}{OVT-OpenCLIP}} ViT-B/32       & OpenCLIP-ViT-B/32                  & LAION (2B)                         &     151M                                                      &   6.6M                                                           & 224        & 512        & 35k       &   67.8~(\textcolor{Green}{$\uparrow$1.3})     &  59.5~(\textcolor{Green}{$\uparrow$22.4})              \\
\textbf{\textcolor{BlueViolet}{OVT-OpenCLIP}} ViT-B/16       & OpenCLIP-ViT-B/16& LAION (2B)& 149M                                                      & 6.6M                                                         & 224        & 512        & 35k    & 69.7 (\textcolor{Green}{$\uparrow$2.1})   & 61.7 (\textcolor{Green}{$\uparrow$17.5})       \\
\textbf{\textcolor{BlueViolet}{OVT-OpenCLIP}} ViT-L/14       & OpenCLIP-ViT-L/14                  & LAION (2B)                         &     428M                                                      &    11.8M                                                          & 224        & 256        &  20k      & 77.3 (\textcolor{Green}{$\uparrow$ 2.1})    & 69.8 (\textcolor{Green}{$\uparrow$16.6})     \\ \hline
\textbf{\textcolor{violet}{OVT-MetaCLIP}} ViT-B/32       & MetaCLIP-ViT-B/32                   & Common Crawl (2.5B)                &     151M                                                      &    6.6M                                                          & 224        & 512           & 40k       & 69.7 (\textcolor{Green}{$\uparrow$2.1})       &  54.8 (\textcolor{Green}{$\uparrow$13.8})               \\
\textbf{\textcolor{violet}{OVT-MetaCLIP}} ViT-B/16       & MetaCLIP-ViT-B/16                   & Common Crawl (2.5B)                & 149M                                                      &    6.6M                                                          & 224        &  512          &  40k      & 73.8 (\textcolor{Green}{$\uparrow$1.7})       & 64.8 (\textcolor{Green}{$\uparrow$15.2})                \\
\textbf{\textcolor{violet}{OVT-MetaCLIP}} ViT-L/14       & MetaCLIP-ViT-L/14                   & Common Crawl (2.5B)                &      428M                                                      &    11.8M                                                          & 224        &  256          & 20k       & 77.7 (\textcolor{Maroon}{$\downarrow$1.4})       &  75.4 (\textcolor{Green}{$\uparrow$9.0})               \\ \hline
OVT-BLIP ViT-B/16          & Salesforce-BLIP-ViT-B/16            & COCO~\etal (129M)                        &   234M                                                        & 4.3M                                                             & 224        & 256            &    20k    & 61.7 (\textcolor{Green}{$\uparrow$8.8})       & 54.8 (\textcolor{Green}{$\uparrow$18.0})               \\ \hline
\end{tabular}}
\label{exp:0}
\vspace{-0.3cm}
\end{table*}
\begin{figure}[htb]
  \centering
  \includegraphics[height=3.7cm]{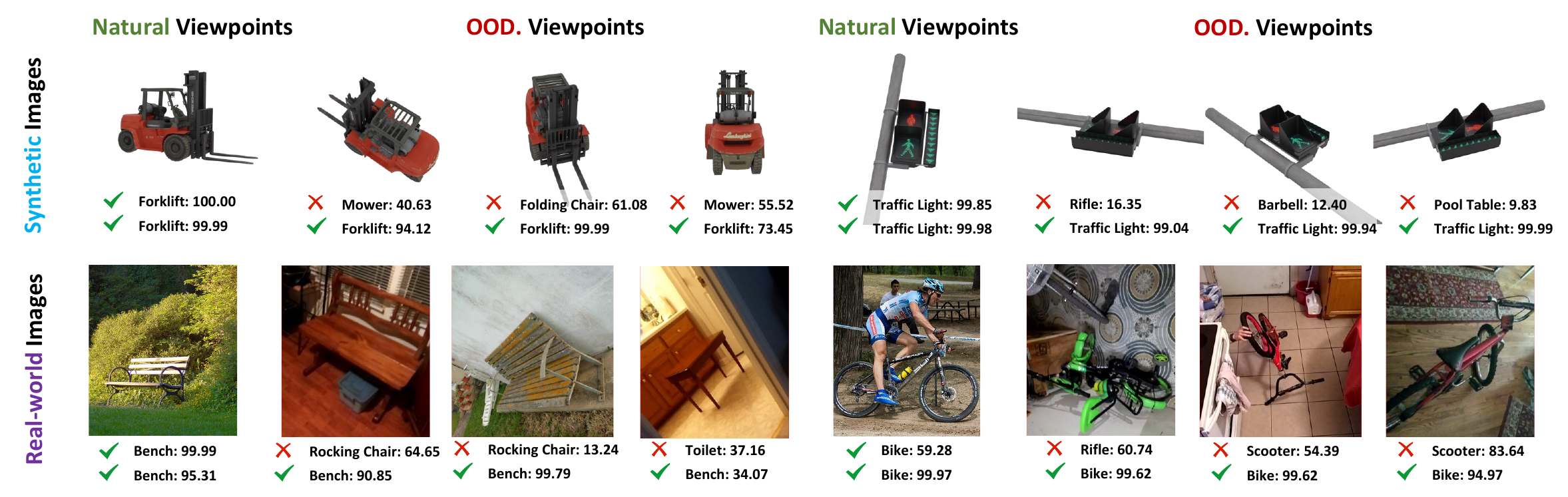}
    \vspace{-0.2cm}
  \caption{\textbf{Visualization for zero-shot classification results. } We select viewpoint-OOD samples of synthetic and real-world scenarios. Below each image, we show the predicted categories and their confidence levels (\%) by the OpenCLIP(ViT-B/16) (\emph{first column}) and by our improved OVT-OpenCLIP(ViT-B/16) (\emph{second column}). \includegraphics[height=0.5em]{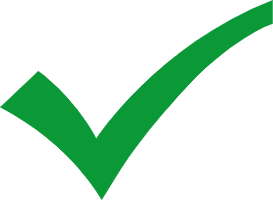} indicates a correct prediction while \includegraphics[height=0.5em]{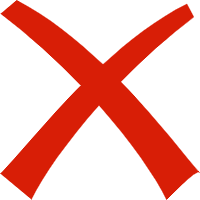} indicating an incorrect one.
  }
  \label{fig:vis3}
  \vspace{-0.5cm}
\end{figure}

\section{Experiments}
Our evaluation of Omniview-Tuning spans several downstream tasks, including zero-shot classification, image captioning, and vision question answering. For zero-shot classification (\cref{sec:exp zeroshot}), we conduct evaluations for CLIP~\cite{radford2021learning} and BLIP~\cite{li2022blip} architectures. For image captioning and vision question answering (\cref{sec:exp vlm}), we replace the visual encoders in Vision Large Language Models (VLLMs) with our fine-tuned versions. We adopt LLaVA-1.5~\cite{liu2024visual,liu2023improved}, and OpenFlamingo~\cite{awadalla2023openflamingo}, the most advanced open-source VLLMs available. Additionally, we present the ablation study and convergence analysis of our approach in \cref{sec:ablation}.

\begin{table*}[t]
\caption{Top-1/Top-5 zero-shot accuracy (\%) under different benchmarks}
  \vspace{-0.2cm}
\setlength\tabcolsep{1.5pt}
\renewcommand\arraystretch{1.5}
\centering
\resizebox{\textwidth}{!}{\begin{tabular}{rcccccccccccccccc}
\hline
\multicolumn{1}{r|}{}               & \multicolumn{4}{c|}{\textbf{Clean}}                                               & \multicolumn{6}{c|}{\textbf{Common-OOD}}                                                                           & \multicolumn{5}{c|}{\textbf{Viewpoint-OOD}}                                                           &                            \\
\multicolumn{1}{c|}{\textbf{Model}} & \rotatebox{90}{ImageNet-100~\cite{ruan2023towards}} & \rotatebox{90}{ImageNet-1K~\cite{deng2009imagenet}} & \rotatebox{90}{Cifar-100~\cite{krizhevsky2009learning}} & \multicolumn{1}{c|}{\textbf{\rotatebox{90}{Avg. Acc.}}} & \rotatebox{90}{ImageNet-V2~\cite{recht2019imagenet}} & \rotatebox{90}{ImageNet-Ske.~\cite{wang2019learning}} & \rotatebox{90}{ImageNet-OOD.~\cite{hendrycks2021natural}} & \rotatebox{90}{ImageNet-Ren.~\cite{hendrycks2021many}} & \rotatebox{90}{OOD-CV~\cite{zhao2022ood}}    & \multicolumn{1}{c|}{\textbf{\rotatebox{90}{Avg. Acc.}}} & \rotatebox{90}{ImageNet-View.~\cite{dong2022viewfool}} & \rotatebox{90}{ImageNet-View.+~\cite{ruan2023towards}} & \rotatebox{90}{OOD-CV-Pose~\cite{zhao2022ood}} & \rotatebox{90}{MIRO~\cite{cha2022miro}}      & \multicolumn{1}{c|}{\textbf{\rotatebox{90}{Avg. Top-1}}} & \textbf{\rotatebox{90}{Total Avg. Acc.}} \\ \hline
\multicolumn{17}{c}{A. Comparisons with   ViT-B/32 baselines}                                                                                                                                                                                                                                                                                                                     \\ \hline
\multicolumn{1}{r|}{\scriptsize OpenAI CLIP}    & 77.5/93.9    & 63.3/88.8   & 64.3/88.1 & \multicolumn{1}{c|}{68.4/90.2}                & 55.8/83.4   & 42.2/70.3     & 33.4/62.2     & 50.7/75.4     & 50.2/82.6 & \multicolumn{1}{c|}{46.5/74.8}                & 44.5/65.4      & 27.5/52.4       & 47.2/84.5   & 26.5/59.4 & \multicolumn{1}{c|}{36.4/65.4}                & 48.6/75.5                       \\
\multicolumn{1}{r|}{\scriptsize Open CLIP}      & \textbf{81.1}/95.3    & 66.5/89.9   & \textbf{75.8}/\textbf{94.0} & \multicolumn{1}{c|}{\textbf{74.5/93.0}}       & \textbf{58.1}/83.9   & \textbf{53.6}/\textbf{79.3}     & 34.8/64.4     & \textbf{61.0}/\textbf{81.9}     & \textbf{53.5}/\textbf{81.9} & \multicolumn{1}{c|}{\textbf{52.2/78.3}}       & 54.4/72.1      & 37.1/63.2       & 46.9/81.6   & 33.0/69.2 & \multicolumn{1}{c|}{42.8/71.5}                & 54.6/79.7                      \\
\rowcolor{BlueViolet!10}\multicolumn{1}{r|}{\scriptsize \textbf{\textcolor{BlueViolet}{OVT-OpenCLIP}}}   & 80.9/95.6    & 67.8/90.8   & 65.0/89.3 & \multicolumn{1}{c|}{71.2/91.9}                & 58.0/84.2   & 45.8/73.4     & 42.8/75.0     & 50.3/71.4     & 51.7/79.5 & \multicolumn{1}{c|}{49.7/76.7}                & \textbf{61.9}/\textbf{81.2}      & \textbf{59.5}/\textbf{85.6}       & \textbf{52.8}/\textbf{82.5}   & \textbf{35.4}/\textbf{80.1} & \multicolumn{1}{c|}{\textbf{52.4/82.4}}       & \textbf{56.0/82.4}              \\ \hline
\multicolumn{1}{r|}{\scriptsize MetaCLIP}       & 80.7/95.6    & 67.6/90.5   & \textbf{77.7}/\textbf{95.2} & \multicolumn{1}{c|}{\textbf{75.3/93.8}}       & 59.5/85.4   & \textbf{55.9}/\textbf{81.4}     & 32.4/62.5     & \textbf{63.2}/\textbf{83.8}     & \textbf{52.0}/\textbf{84.2} & \multicolumn{1}{c|}{\textbf{52.6/79.5}}       & 61.4/76.7      & 41.0/67.8       & 48.9/\textbf{87.9}   & 34.8/73.2 & \multicolumn{1}{c|}{46.5/76.4}                & 56.3/82.0                       \\
\rowcolor{violet!10}\multicolumn{1}{r|}{\scriptsize \textbf{\textcolor{violet}{OVT-MetaCLIP}}}    & \textbf{80.7}/\textbf{95.6}    & \textbf{69.7}/\textbf{92.0}   & 71.8/93.0 & \multicolumn{1}{c|}{74.0/93.5}                & \textbf{60.6}/\textbf{85.8}   & 47.8/75.8     & \textbf{43.5}/\textbf{73.8}     & 49.0/70.8     & 50.1/80.1 & \multicolumn{1}{c|}{50.2/77.2}                & \textbf{64.0}/\textbf{79.2}      & \textbf{54.8}/\textbf{80.4}       & \textbf{55.1}/84.8   & \textbf{35.6}/\textbf{77.0} & \multicolumn{1}{c|}{\textbf{52.4/80.3}}       & \textbf{56.9/82.3}              \\ \hline
\multicolumn{17}{c}{B. Comparisons with   ViT-B/16 baselines}                                                                                                                                                                                                                                                                                                                     \\ \hline
\multicolumn{1}{r|}{\scriptsize OpenAI CLIP}    & 82.1/95.7    & 68.3/91.9   & 67.2/89.4 & \multicolumn{1}{c|}{72.5/92.3}                & 61.8/87.4   & 48.2/76.3     & 27.7/55.7     & 59.1/83.0     & 52.2/\textbf{84.6} & \multicolumn{1}{c|}{49.8/77.4}                & 51.6/68.9      & 36.9/63.8       & 53.4/86.8   & 30.1/66.1 & \multicolumn{1}{c|}{43.0/71.4}                & 53.2/79.1                       \\
\multicolumn{1}{r|}{\scriptsize Open CLIP}      & 83.2/96.2    & 70.1/91.8   & \textbf{77.0}/\textbf{94.8} & \multicolumn{1}{c|}{\textbf{76.8/94.3}}       & 62.2/87.0   & \textbf{56.0}/\textbf{82.0}     & 30.7/59.8     & \textbf{64.9}/\textbf{85.6}     & 54.3/82.7 & \multicolumn{1}{c|}{\textbf{53.6/79.4}}       & 58.1/74.4      & 44.2/70.9       & 48.5/84.0   & 34.6/74.6 & \multicolumn{1}{c|}{46.4/76.0}                & 57.0/82.0                       \\
\rowcolor{BlueViolet!10}\multicolumn{1}{r|}{\scriptsize \textbf{\textcolor{BlueViolet}{OVT-OpenCLIP}}}   & \textbf{83.9}/\textbf{97.0}    & \textbf{71.9}/\textbf{93.1}   & 69.0/90.7 & \multicolumn{1}{c|}{74.9/93.6}                & \textbf{64.0}/\textbf{88.6}   & 50.5/77.9     & \textbf{36.8}/\textbf{68.9}     & 57.0/77.2     & \textbf{56.3}/84.5 & \multicolumn{1}{c|}{52.9/\textbf{79.4}}                & \textbf{65.4}/\textbf{80.7}      & \textbf{61.7}/\textbf{85.8}       & \textbf{56.9}/\textbf{87.4}   & \textbf{42.4}/\textbf{84.9} & \multicolumn{1}{c|}{\textbf{56.6/84.7}}       & \textbf{59.6/84.7}              \\ \hline
\multicolumn{1}{r|}{\scriptsize EVA-CLIP}       & \textbf{85.3}/96.5    & \textbf{74.6}/\textbf{94.2}   & 87.5/\textbf{98.0} & \multicolumn{1}{c|}{\textbf{82.5/96.3}}       & \textbf{67.0}/\textbf{89.8}   & 57.6/82.3     & 21.3/47.3     & 69.6/87.5     & \textbf{53.1}/83.1 & \multicolumn{1}{c|}{53.7/78.0}                & 61.8/76.6      & 44.3/69.4       & 53.9/87.4   & 32.9/73.2 & \multicolumn{1}{c|}{48.2/76.6}                & 59.1/82.1                       \\
\multicolumn{1}{r|}{\scriptsize MetaCLIP}       & 84.3/97.2    & 72.1/93.4   & \textbf{78.9}/95.4 & \multicolumn{1}{c|}{78.4/95.3}                & 65.0/89.3   & \textbf{60.1}/\textbf{84.8}     & 26.2/56.4     & \textbf{70.2}/\textbf{89.3}     & 52.3/85.4 & \multicolumn{1}{c|}{\textbf{54.8/81.0}}       & 64.2/79.4      & 49.6/76.1       & 48.9/\textbf{90.9}   & 38.5/78.7 & \multicolumn{1}{c|}{50.3/81.2}                & 59.2/84.7                       \\
\rowcolor{violet!10}\multicolumn{1}{r|}{\scriptsize \textbf{\textcolor{violet}{OVT-MetaCLIP}}}    & 83.4/\textbf{97.4}    & 73.8/94.1   & 73.9/93.6 & \multicolumn{1}{c|}{77.0/95.0}                & 65.9/89.4   & 53.6/81.0     & \textbf{36.2}/\textbf{66.8}     & 59.0/79.6     & 51.6/\textbf{83.8 }& \multicolumn{1}{c|}{53.2/80.1}                & \textbf{69.7}/\textbf{84.0}      & \textbf{64.8}/\textbf{87.3}       & \textbf{55.2}/87.8   & \textbf{39.2}/\textbf{82.9} & \multicolumn{1}{c|}{\textbf{57.2/85.5}}       & \textbf{60.5/85.6}              \\ \hline
\multicolumn{17}{c}{C. Comparisons with   ViT-L/14 baselines}                                                                                                                                                                                                                                                                                                                     \\ \hline
\multicolumn{1}{r|}{\scriptsize OpenAI CLIP}    & 86.5/97.4    & 75.4/94.6   & 76.5/93.3 & \multicolumn{1}{c|}{79.5/95.1}                & \textbf{69.8}/90.9   & 59.5/84.3     & 18.6/43.8     & 72.8/91.4     & 52.9/88.8 & \multicolumn{1}{c|}{54.7/79.8}                & 60.3/75.6      & 45.8/71.5       & 47.9/88.2   & 38.0/74.1 & \multicolumn{1}{c|}{48.0/77.3}                & 58.6/82.8                       \\
\multicolumn{1}{r|}{\scriptsize Open CLIP}      & 86.8/97.8    & 75.2/94.3   & \textbf{83.7}/\textbf{96.7} & \multicolumn{1}{c|}{\textbf{81.9/96.2}}       & 67.7/90.2   & \textbf{63.2}/\textbf{86.4}     & 24.0/50.5     & \textbf{74.5}/\textbf{91.2}     & 54.5/85.0 & \multicolumn{1}{c|}{56.8/80.6}                & 65.7/78.1      & 53.2/76.7       & 52.4/90.5   & 42.3/83.0 & \multicolumn{1}{c|}{53.4/82.1}                & 61.9/85.0                       \\
\rowcolor{BlueViolet!10}\multicolumn{1}{r|}{\scriptsize \textbf{\textcolor{BlueViolet}{OVT-OpenCLIP}}}   & \textbf{89.0}/\textbf{97.8}    & \textbf{77.3}/\textbf{95.3}   & 79.2/95.3 & \multicolumn{1}{c|}{81.8/96.1}                & 69.6/\textbf{91.5}   & 61.9/86.0     & \textbf{27.5}/\textbf{55.4}     & 71.3/88.7     & \textbf{56.4}/\textbf{87.0} & \multicolumn{1}{c|}{\textbf{57.3/81.7}}       & \textbf{72.2}/\textbf{86.6}      & \textbf{69.8}/\textbf{89.7}       & \textbf{57.3}/\textbf{94.1}   & \textbf{50.0}/\textbf{89.3} & \multicolumn{1}{c|}{\textbf{62.3/89.9}}       & \textbf{65.1/88.1}              \\ \hline
\multicolumn{1}{r|}{\scriptsize EVA-CLIP}       & 88.5/97.9    & \textbf{79.6}/\textbf{96.0}   & \textbf{90.6}/\textbf{98.6 }& \multicolumn{1}{c|}{\textbf{86.3/97.5}}       & \textbf{72.8}/\textbf{92.7}   & 68.0/89.1     & 16.3/40.0     & \textbf{82.8}/\textbf{95.7}     & 54.7/87.4 & \multicolumn{1}{c|}{58.9/81.0}                & 71.5/82.3      & 61.1/81.7       & 54.4/\textbf{94.5}   & 39.6/86.1 & \multicolumn{1}{c|}{56.6/86.1}                & 65.0/86.8                       \\
\multicolumn{1}{r|}{\scriptsize MetaCLIP}       & 88.3/\textbf{98.3}    & 79.1/95.9   & 84.1/96.9 & \multicolumn{1}{c|}{83.8/97.0}                & 72.5/92.6   & \textbf{68.9}/\textbf{89.8}     & 17.0/40.6     & 81.8/95.1     & \textbf{56.6}/87.5 & \multicolumn{1}{c|}{\textbf{59.3}/81.1}       & 77.3/89.3      & 66.4/87.0       & \textbf{58.9}/93.3   & \textbf{48.1}/89.6 & \multicolumn{1}{c|}{62.7/89.8}                & \textbf{66.6}/88.0              \\
\rowcolor{violet!10}\multicolumn{1}{r|}{\scriptsize \textbf{\textcolor{violet}{OVT-MetaCLIP}}}   & \textbf{88.8}/97.5    & 77.7/95.9   & 84.0/96.9 & \multicolumn{1}{c|}{83.5/96.8}                & 70.8/92.2   & 64.4/87.9     & \textbf{20.8}/\textbf{47.0}     & 77.0/92.7     & 56.3/\textbf{89.3} & \multicolumn{1}{c|}{57.8/\textbf{81.8}}                & \textbf{79.3}/\textbf{90.6}      & \textbf{75.4}/\textbf{93.0}       & 57.0/94.4   & 46.4/\textbf{93.8} & \multicolumn{1}{c|}{\textbf{64.5/92.9}}       & 66.5/\textbf{89.3}                      \\ \hline
\multicolumn{17}{c}{D. Comparisons with   BLIP ViT-B/16 baselines}                                                                                                                                                                                                                                                                                                                \\ \hline
\multicolumn{1}{r|}{\scriptsize BLIP}           & 76.6/93.3    & 52.9/80.2   & \textbf{67.0}/88.3 & \multicolumn{1}{c|}{65.5/87.3}                & 47.3/74.7   & \textbf{51.0}/\textbf{76.6}     & 25.6/53.4     & \textbf{64.3}/\textbf{83.8}     & 53.9/\textbf{87.6} & \multicolumn{1}{c|}{48.4/75.2}                & 55.2/68.2      & 36.8/63.3       & 50.8/\textbf{89.9}   & 27.0/66.1 & \multicolumn{1}{c|}{42.4/71.9}                & 50.7/77.1                       \\
\rowcolor{gray!25}\multicolumn{1}{r|}{\scriptsize OVT-BLIP}        &\textbf{ 82.2}/\textbf{97.0}    & \textbf{61.7}/\textbf{88.8}   & 66.6/\textbf{88.9} & \multicolumn{1}{c|}{\textbf{70.2/91.5}}       & \textbf{53.7}/\textbf{82.9}   & 46.5/74.2     &\textbf{ 33.8}/\textbf{62.7}     & 57.4/77.9     & \textbf{56.4}/87.3 & \multicolumn{1}{c|}{\textbf{49.6/77.0}}       & \textbf{62.6}/\textbf{79.0}      & \textbf{54.8}/\textbf{79.9}       & \textbf{55.2}/89.5   & \textbf{31.5}/\textbf{73.2} & \multicolumn{1}{c|}{\textbf{51.0/80.4}}       & \textbf{55.2/81.8}              \\ \hline
\end{tabular}}
  \vspace{-0.3cm}
\label{exp:1}
\end{table*}

\subsection{Evaluation of Zero-Shot Classification} \label{sec:exp zeroshot} 

\noindent \textbf{Baselines.} We adopt the official CLIP (OpenAI CLIP~\cite{radford2021learning}) and the community open-source version (OpenCLIP~\cite{ilharco_gabriel_2021_5143773}) as our baselines. Additionally, we include the current state-of-the-art Eva02-CLIP~\cite{sun2023eva} and MetaCLIP~\cite{xu2023demystifying} as another set of baselines to compare CLIP versions trained with improved techniques and more extensive training data. For BLIP, we use the official implementation~\cite{li2022blip} as the baseline. All models are evaluated using publicly available weights.

\noindent \textbf{Settings.} We train two series of CLIP models using our OVT framework and MVCap dataset, each series comprising three different visual encoder architectures (ViT-B/32, ViT-B/16, and ViT-L/14). \textcolor{BlueViolet}{OVT-OpenCLIP} are fine-tuned on the original weights of OpenCLIP, while \textcolor{violet}{OVT-MetaCLIP} are based on the weights of MetaCLIP. The fine-tuning settings for each OVT model are detailed in~\cref{exp:0}. We standardized the $\lambda$=1.0, $\alpha$=0.1, and the number of outlier viewpoints $K$=5 and set the rank of LoRA at 8. The ablation results for key hyperparameters will be reported in~\cref{sec:ablation}.

\noindent \textbf{Datasets and Metrics.}  We employ a various set of benchmarks for evaluation, including clean data distributions (ImageNet~\cite{deng2009imagenet} and CIFAR~\cite{krizhevsky2009learning}), common 2D-OOD (ImageNet-V2~\cite{recht2019imagenet}, ImageNet-Sketch~\cite{wang2019learning}, ImageNet-O~\cite{hendrycks2021natural}, ImageNet-R~\cite{hendrycks2021many} and OOD-CV~\cite{zhao2022ood}), and most importantly, viewpoint-OOD (ImageNet-V~\cite{dong2022viewfool}, ImageNet-V+~\cite{ruan2023towards}, OOD-CV(Pose)~\cite{zhao2022ood} and MIRO~\cite{cha2022miro}) datasets. For each benchmark, we report Top-1 and Top-5 accuracy and average accuracy across all benchmarks. The evaluations follow the standard prompting engineering and candidate category names conventions of CLIP~\cite{radford2021learning}.

\noindent \textbf{Results and Discussions.} 
\cref{exp:1} summarizes the performance of our OVT-trained VLP models against various VLP versions, including their accuracy across different benchmarks and average accuracy across clean, common-OOD, and Viewpoint-OOD domains. We can draw the following conclusions:

\noindent\underline{\textbf{(1)}} OVT significantly enhances the models' invariance to Viewpoint-OOD samples. Across different VLP architectures and visual encoders, OVT-trained models perform best on almost all viewpoint-OOD benchmarks. On the average accuracy of viewpoint-OOD datasets, OVT-OpenCLIP with ViT-B/32, ViT-B/16, and ViT-L/14 shows improvements of 9.6\%, 10.2\%, and 8.9\% over OpenCLIP, respectively. OVT-BLIP demonstrated an average improvement of 8.6\%. 

\noindent\underline{\textbf{(2)}}  While enhancing viewpoint invariance, OVT maintains good performance on clean samples and 2D-OOD without significant performance trade-offs. For 2D-OOD benchmarks, OVT-OpenCLIP with ViT-B/32, ViT-B/16, and ViT-L/14 sacrifice only 2.6\%, 1.4\%, and 0.2\% accuracy. 

\noindent\underline{\textbf{(3)}}  Compared to earlier CLIP baselines, the recently developed MetaCLIP exhibits better zero-shot performance and robustness. Based on this, OVT further enhances its performance under viewpoint-OOD samples. 

\noindent \textbf{Visualization.}
We showcase OVT-OpenCLIP and the original OpenCLIP prediction on several viewpoint-OOD samples. As illustrated in~\cref{fig:vis1}, OVT-CLIP successfully predicts the categories of images from various unusual viewpoints in all cases, whereas the original CLIP is prone to make incorrect predictions.

\begin{table*}[t]
\caption{Image captioning performance under clean distribution samples and viewpoint-OOD samples from Real-world and Synthetic domains. We utilize the MPNet~\cite{song2020mpnet} to calculate the similarity between generated descriptions and ground-truth labels, considering predictions successful if they exceed the similarity threshold $\beta$.}

\setlength\tabcolsep{2.3pt}
\renewcommand\arraystretch{1.2}
\centering
\resizebox{\textwidth}{!}{\begin{tabular}{c|r|cccccc|cccccc}
\hline
                               &                                              & \multicolumn{6}{c|}{\textbf{Real-world Domain}}                                                    & \multicolumn{6}{c}{\textbf{Synthetic Domain}}                                                       \\ \cline{3-14} 
                               &                                              & \multicolumn{3}{c|}{OOD-CV (iid)~\cite{zhao2022ood}}                    & \multicolumn{3}{c|}{OOD-CV (Pose)~\cite{zhao2022ood}} & \multicolumn{3}{c|}{IM3D~\cite{ruan2023towards}}                            & \multicolumn{3}{c}{ImageNet-V+~\cite{ruan2023towards}} \\
\textbf{Model}                 & \multicolumn{1}{c|}{\textbf{Visual Encoder}} & $\beta@1.0$ & $\beta@0.5$ & \multicolumn{1}{c|}{$\beta@Adp.$} & $\beta@1.0$  & $\beta@0.5$  & $\beta@Adp.$  & $\beta@1.0$ & $\beta@0.5$ & \multicolumn{1}{c|}{$\beta@Adp.$} & $\beta@1.0$   & $\beta@0.5$  & $\beta@Adp.$  \\ \hline \hline
\multirow{4}{*}{LLaVa-7b}  & \scriptsize OpenAI CLIP(ViT-L/14)                        & 44.1         & 61.1         & \multicolumn{1}{c|}{67.5}          & 46.4          & \textbf{53.6}          & 58.7           & 46.7         & 53.3         & \multicolumn{1}{c|}{58.8}          & 20.4           & 25.5          & 32.1           \\
                               & \scriptsize $TeCoA^{4}$~\cite{mao2022understanding}(ViT-L/14)           & 41.9    & 58.9    & \multicolumn{1}{c|}{65.5}     & 36.1     & 41.6     & 49.2       & 26.3    & 30.1    & \multicolumn{1}{c|}{42.6}     & 8.7       & 11.6     & 22.6      \\
                               & \scriptsize $FARE^{4}$~\cite{schlarmann2024robust}(ViT-L/14)            & 42.1    & 58.9    & \multicolumn{1}{c|}{65.2}      & 40.2     & 45.9     & 50.8      & 35.2     & 39.2    & \multicolumn{1}{c|}{49.2}     & 12.7      & 15.8     & 23.1      \\
                               \rowcolor{gray!25}& \scriptsize OVT-CLIP(ViT-L/14)                           & 43.5         & 59.5         & \multicolumn{1}{c|}{65.9}          & \textbf{46.5}          & \textbf{53.6}          & \textbf{59.1}           & 49.4         & 54.0& \multicolumn{1}{c|}{61.8}          & \textbf{26.4}           & \textbf{31.9}          & \textbf{41.0}\\ \hline
\multirow{4}{*}{LLaVa-13b} & \scriptsize OpenAI CLIP(ViT-L/14)                        & 45.4         & 68.0& \multicolumn{1}{c|}{70.6}          & \textbf{48.6}          &\textbf{58.6}          & 60.8           & 48.7         & 56.7         & \multicolumn{1}{c|}{60.8}          & 21.2           & 28.4          & 32.5           \\
                               & \scriptsize $TeCoA^{4}$~\cite{mao2022understanding}(ViT-L/14)           & 42.4         & 67.0& \multicolumn{1}{c|}{72.2}          & 37.4          & 48.9          & 51.3           & 25.0& 28.6         & \multicolumn{1}{c|}{41.5}          & 8.4           & 10.9          & 21.8           \\
                               & \scriptsize $FARE^{4}$~\cite{schlarmann2024robust}(ViT-L/14)            & 43.9    & 66.7     & \multicolumn{1}{c|}{71.1}     & 41.9     & 52.1     & 54.8      & 36.1     & 41.4    & \multicolumn{1}{c|}{48.6}     & 12.1      & 15.9     & 20.8      \\
                               \rowcolor{gray!25}& \scriptsize OVT-CLIP(ViT-L/14)                           & 45.7         & 67.3         & \multicolumn{1}{c|}{70.8}          & 48.2          & \textbf{58.6}          & \textbf{61.9}           & 50.4         & 58.9         & \multicolumn{1}{c|}{63.2}          & \textbf{26.4}           & \textbf{36.2}          & \textbf{40.9}           \\ \hline
\end{tabular}}
  \vspace{-0.2cm}
\label{exp:3}
\end{table*}


\begin{figure}[t]
  \centering 
  \begin{minipage}[t]{0.65\textwidth}
    \centering
    \includegraphics[width=\linewidth]{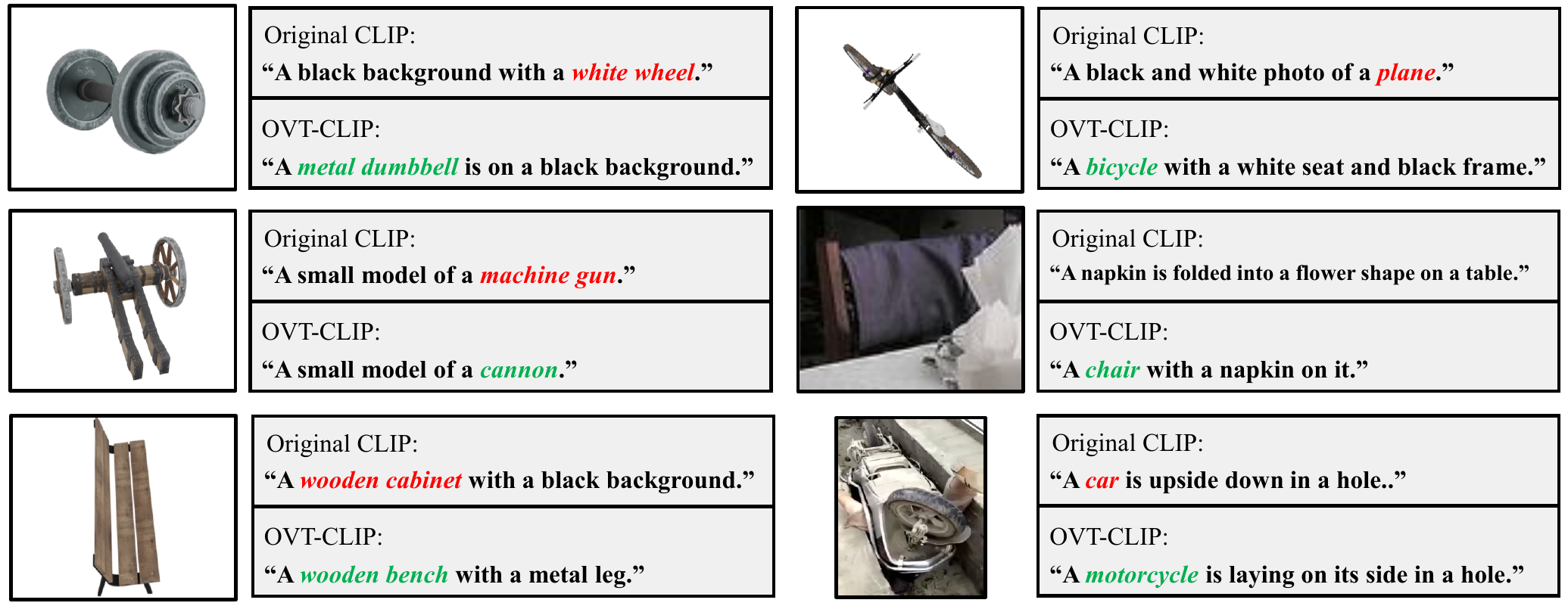}
    \caption{The image descriptions generated by LLaVa-13B using our OVT-CLIP and the original OpenAI CLIP as vision encoder, where \textcolor{red}{\emph{\textbf{red texts}}} indicates incorrect category descriptions, and \textcolor{Green}{\emph{\textbf{green texts}}} represents correct.}
    \label{fig:vis4}
  \end{minipage}
  \hfill 
  \begin{minipage}[t]{0.30\textwidth}
    \centering
    \includegraphics[width=\linewidth]{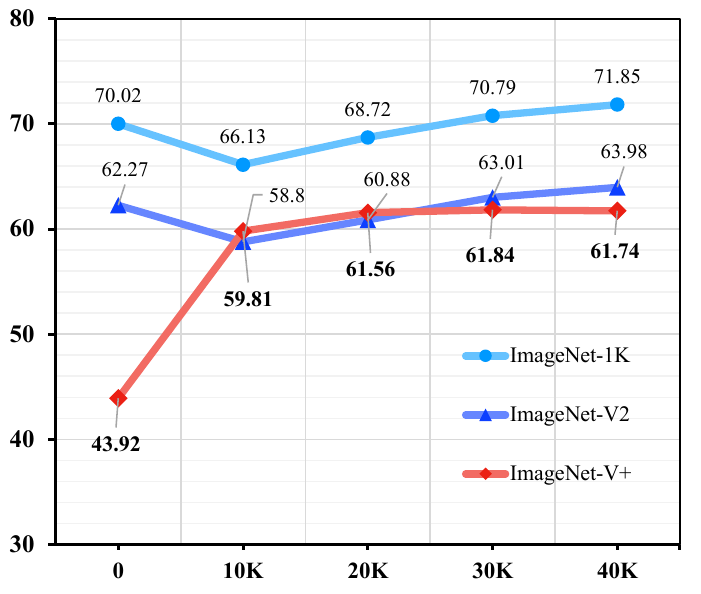}
    \caption{The Top-1 accuracy of OVT-OpenCLIP (ViT-B/16) with the iterations increases.}
    \label{fig:vis5}
  \end{minipage}
    \vspace{-0.5cm}
\end{figure}

\subsection{Performance on Other Tasks} \label{sec:exp vlm}
\noindent \textbf{Settings.} As LLaVA and Openflamingo use the OpenAI CLIP (ViT-L/14) to encode vision inputs, we applied OVT to this model in this section and comparing with other OpenAI CLIP (ViT-L/14) versions in image captioning tasks. The training setup remains consistent with the OVT-OpenCLIP described in~\cref{exp:0}. 

\noindent \textbf{Baselines.} In addition to comparing with the original OpenAI CLIP version, we also select $TeCoA^4$~\cite{mao2022understanding} and $FARE^4$~\cite{schlarmann2024robust}, robust CLIP models based on adversarial training, as baselines, which have been proven to possess good resistance to adversarial samples in image captioning task.

\noindent \textbf{Datasets and Metrics.} Given the absence of caption benchmarks that include viewpoint-changing OOD samples, we conduct evaluations using existing viewpoint-OOD datasets, including real-world datasets (using OOD-CV (iid) to represent clean distribution and OOD-CV (Pose) for viewpoint-OOD) and synthetic datasets (using IM3D~\cite{ruan2023towards} for clean distribution and ImageNet-V+ for viewpoint-OOD). We adopt word embedding distance to calculate the accuracy of the captioning task. By adopting MPNet~\cite{song2020mpnet}, a state-of-the-art textual embedding model, we measure the similarity between keywords in the generated description and the ground-truth categories. Then assess the accuracy by counting the number of samples that exceed a specific similarity threshold $\beta$. 

\noindent \textbf{Results and Discussions.} 
\cref{exp:3} shows the image captioning accuracy of CLIP models under different training strategies, considering $\beta$ at 1.0 (indicating predictions involve ground-truth categories), 0.5, and $Adp.$ (meaning $\beta$ is equal to the average similarity in the clean distribution). We found that OVT-CLIP improves the accuracy of descriptions generated by LLaVa for viewpoint-OOD samples while maintaining its performance on corresponding clean distributions. When used as the visual encoder for the LLaVa-7B model, OVT-CLIP achieved an 8.9\% increase in accuracy compared to the original CLIP model weights under $\beta=Adp$. Besides, we find that although robust CLIP versions maintain performance on clean distribution samples, they experience a significant performance decline when facing viewpoint-OOD samples. We select some examples with the generated description in~\cref{fig:vis4}. For the visual question-answering task, we used OpenFlamingo as the VLLMs. The results are reported in the Appendix~\textcolor{red}{A}.

\subsection{Ablation Studies and Additional Results} \label{sec:ablation}

\begin{table*}[t]
\caption{Average Top-1/Top-5 zero-shot accuracy (\%) under different data distributions within various ablation settings.}
  \vspace{-0.2cm}
\setlength\tabcolsep{2.3pt}
\renewcommand\arraystretch{1.3}
\centering
\resizebox{\textwidth}{!}{\begin{tabular}{ccc|cc|cc|cc|cc}
\hline
\textbf{$\mathcal{L}_{ITC}$} & VIFormer & \textbf{$\mathcal{L}_{VC}$} & \multicolumn{2}{c|}{\textbf{Total Avg.}}                    & \multicolumn{2}{c|}{\textbf{Clean Avg.}} & \multicolumn{2}{c|}{\textbf{Common-OOD Avg.}} & \multicolumn{2}{c}{\textbf{Viewpoint-OOD Avg.}} \\ \hline
\XSolidBrush             & \XSolidBrush                 & \XSolidBrush            & 61.9                          & 85.0                         & 81.9                & 96.2               & 56.8                  & 80.6                  & 53.4                   & 82.1                   \\
\Checkmark             & \XSolidBrush                 & \XSolidBrush            & 61.9                          & 85.7 (\textcolor{Green}{$\uparrow$0.7})         & 79.4 (\textcolor{Maroon}{$\downarrow$2.5})                & 95.6 (\textcolor{Maroon}{$\downarrow$0.6})                & 56.2 (\textcolor{Maroon}{$\downarrow$0.6})                  & 81.4 (\textcolor{Green}{$\uparrow$0.8})                  & 56.0 (\textcolor{Green}{$\uparrow$2.6})                   & 83.8 (\textcolor{Green}{$\uparrow$1.7})                   \\
\Checkmark             & \Checkmark                 & \XSolidBrush            & 62.2 (\textcolor{Green}{$\uparrow$0.3})          & 86.2 (\textcolor{Green}{$\uparrow$1.2})         & 79.9 (\textcolor{Maroon}{$\downarrow$2.0})                & 95.4 (\textcolor{Maroon}{$\downarrow$0.8})              & 55.4 (\textcolor{Maroon}{$\downarrow$1.4})                  & 81.6 (\textcolor{Green}{$\uparrow$1.0})                  & 57.5 (\textcolor{Green}{$\uparrow$4.1})                   & 85.2 (\textcolor{Green}{$\uparrow$3.1})                   \\
\Checkmark             & \Checkmark                 & \Checkmark            & \textbf{65.1 (\textcolor{Green}{$\uparrow$3.2)}} & \textbf{88.1 (\textcolor{Green}{$\uparrow$3.1)}} & \textbf{81.8 (\textcolor{Maroon}{$\downarrow$0.1)}}       & \textbf{96.1 (\textcolor{Maroon}{$\downarrow$0.1)}}      & \textbf{57.3 (\textcolor{Green}{$\uparrow$0.5)}}         & \textbf{81.7 (\textcolor{Green}{$\uparrow$1.1)}}         & \textbf{62.3 (\textcolor{Green}{$\uparrow$8.9)}}          & \textbf{89.9 (\textcolor{Green}{$\uparrow$7.8)}}          \\ \hline
\end{tabular}}
\label{exp:ablation}
  \vspace{-0.4cm}
\end{table*}

Our ablation studies focus on the VIFormer and the $\mathcal{L}_{VC}$ within the Omniview-Tuning framework. \cref{exp:ablation} shows the Top-1/Top-5 accuracy of OVT-OpenCLIP (ViT-L/14) across various data distributions and different ablation settings. Beyond the original OpenCLIP, we set a baseline that only uses  $\mathcal{L}_{ITC}$ for fine-tuning. Keeping other training settings fixed, reliance solely on $\mathcal{L}_{ITC}$ led to a more significant performance decline in clean and 2D-OOD samples while achieving limited viewpoint OOD performance improvement (2.6\%/1.7\%). The integration of VIFormer led to further improvements in viewpoint OOD accuracy (4.1\%/3.1\%). With the further addition of $\mathcal{L}_{VC}$, the improvement in viewpoint OOD performance is most significant (8.9\%/7.8\%), and it also reduces performance sacrifices in other data distributions. Detailed analyses on the effects of outlier sample count and loss balance parameters are available in Appendix~\textcolor{red}{B}.

Furthermore, we report OVT's training convergence, depicted in \cref{fig:vis5}. We display the Top-1 accuracy evolution for OVT-OpenCLIP (ViT-B/16) across various training iterations. We observe that around 40K iterations, with a batch size of 512, are sufficient for effective convergence, thus achieving a balance in performance across different data distributions.

\section{Conclusions}
To tackle the challenge of 3D viewpoint invariance in VLP models, this paper first introduced the MVCap dataset, a million-scale collection of image-text pairs with diverse viewpoint variations. Building upon this groundwork, we then proposed the Omniview-Tuning framework, which incorporates a novel Cross-Viewpoint Alignment objective in a parameter-efficient manner, effectively enhancing the VLP models' ability to generate viewpoint-invariant representations. Moreover, through extensive experiments, we successfully verified that Omniview-Tuning could bring significant improvements in viewpoint invariance while preserving the original performance. These advancements provide valuable insights and a standard for future research on viewpoint invariance in foundation models.


\clearpage
\appendix

\numberwithin{equation}{section}
\numberwithin{figure}{section}
\numberwithin{table}{section}

\section{Evaluation on OpenFlamingo}
\vspace{-0.2cm}
\begin{figure}[h]
  \centering
  \includegraphics[height=7.8cm]{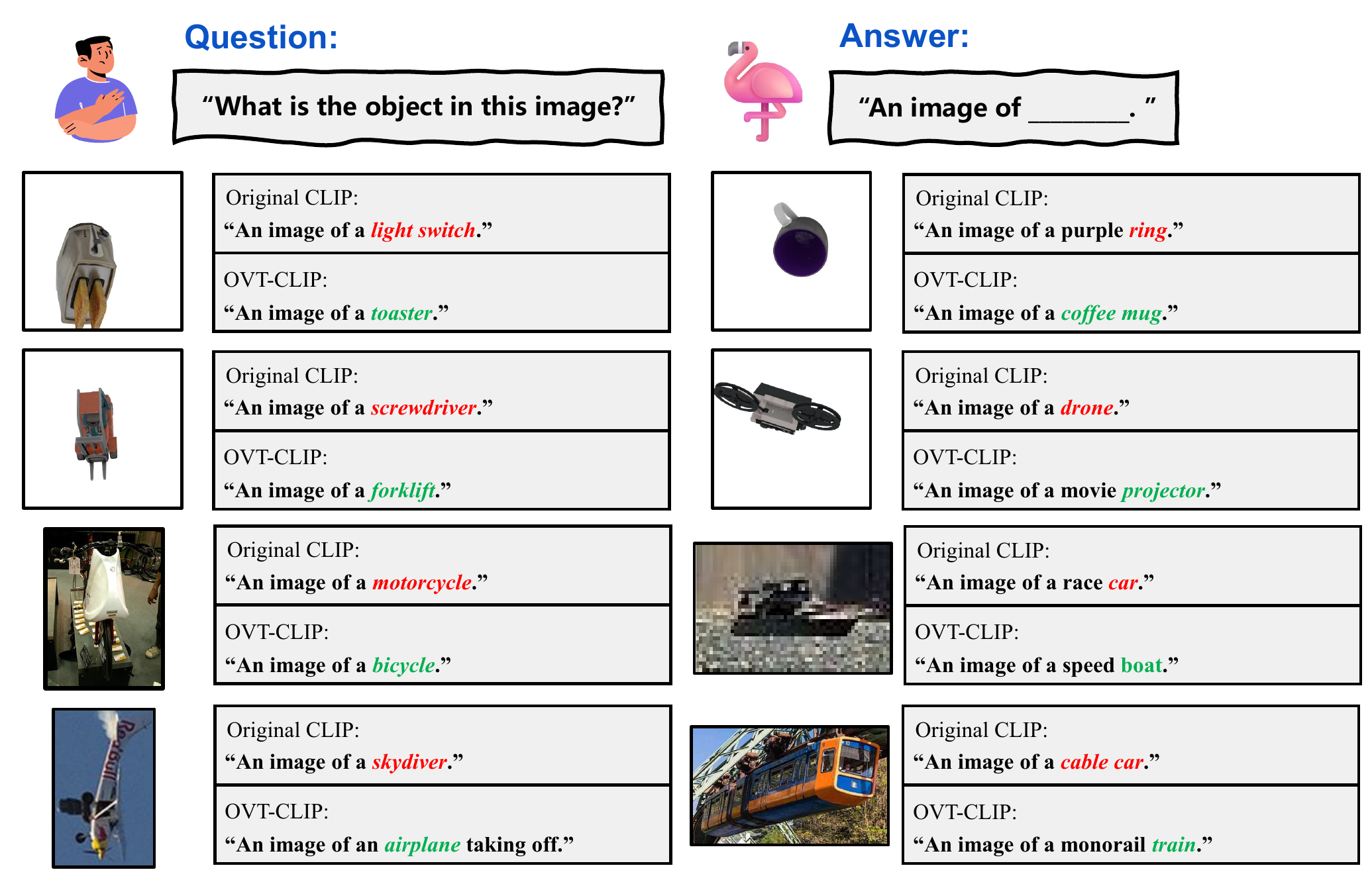}
  \caption{The answers generated by OpenFlamingo-3B using our OVT-CLIP and the original OpenAI CLIP as vision encoder, where \textcolor{red}{\emph{\textbf{red texts}}} indicates incorrect category descriptions, and \textcolor{Green}{\emph{\textbf{green texts}}} represents correct.}
  \label{fig:vis0}
  \vspace{-0.3cm}
\end{figure}
In this study, we integrate our improved OVT-CLIP into OpenFlamingo~\cite{awadalla2023openflamingo} to evaluate its performance in the Visual Question Answering (VQA) task, leveraging the same evaluation datasets and metrics outlined in Sec.~\textcolor{red}{5.2} for consistency. Our experimental setup involves a comparative analysis between the baseline OpenAI CLIP model (ViT-L/14) and our improved OVT-CLIP (ViT-L/14). For OpenFlamingo's text prompts, we employ a question-and-answer format, with the questions template as "What is the object in this image?" and the answers template as "This is an image of <>."

The results across different data distributions are shown in~\cref{exp:vqa}. It indicates that OVT-CLIP significantly outperforms the original OpenAI CLIP model in handling viewpoint-OOD data (OOD-CV(Pose) and ImageNet-V+) while preserving its performance on clean data distributions (OOD-CV(iid) and IM3D) across the 3B and 4B parameter scales in OpenFlamingo. \cref{fig:vis0} highlights specific answer examples where OpenFlamingo, powered by OVT-CLIP, demonstrates remarkable precision in identifying object categories despite shifts in viewpoint. Building on these promising results, we will next focus on extending the application of OVT-CLIP to a broader spectrum of VLLMs to further bolster their resilience against viewpoint shifts, thereby enhancing their overall robustness and applicability in real-world scenarios.

\begin{table*}[t]
\caption{VQA accuracy (\%) of OpenFlamingo under clean distribution samples and viewpoint-OOD samples from Real-world and Synthetic domains. We utilize the MPNet~\cite{song2020mpnet} to calculate the similarity between generated descriptions and ground-truth labels, considering predictions successful if they exceed the similarity threshold $\beta$.}

\setlength\tabcolsep{2.3pt}
\renewcommand\arraystretch{1.2}
\centering
\resizebox{\textwidth}{!}{\begin{tabular}{c|c|cccccc|cccccc}
\hline
                                 &                         & \multicolumn{6}{c|}{\textbf{Real-world Domain}}                                                                                    & \multicolumn{6}{c}{\textbf{Synthetic Domain}}                                                                                      \\ \cline{3-14} 
                                 &                         & \multicolumn{3}{c|}{OOD-CV (iid)}                                  & \multicolumn{3}{c|}{OOD-CV (Pose)}            & \multicolumn{3}{c|}{IM3D}                                          & \multicolumn{3}{c}{ImageNet-View.+}           \\
\textbf{Model}                   & \textbf{Visual Encoder} & $\beta@1.0$      & $\beta@0.5$      & \multicolumn{1}{c|}{$\beta@Adp.$}     & $\beta@1.0$      & $\beta@0.5$      & $\beta@Adp.$     & $\beta@1.0$      & $\beta@0.5$      & \multicolumn{1}{c|}{$\beta@Adp.$}     & $\beta@1.0$      & $\beta@0.5$      & $\beta@Adp.$     \\ \hline \hline
\multirow{2}{*}{OF-3b} & OpenAI CLIP(ViT-L/14)   & 40.1          & \textbf{93.3} & \multicolumn{1}{c|}{62.6}          & 37.7          & \textbf{87.0} & 48.8          & 49.2          & 78.3          & \multicolumn{1}{c|}{59.1}          & 24.6          & 55.8          & 34.2          \\
                                 & OVT-CLIP(ViT-L/14)      & \textbf{40.7} & 92.7          & \multicolumn{1}{c|}{\textbf{63.4}} & \textbf{38.0} & 82.5          & \textbf{49.9} & \textbf{50.7} & \textbf{79.4} & \multicolumn{1}{c|}{\textbf{61.2}} & \textbf{30.0} & \textbf{62.4} & \textbf{42.0} \\ \hline
\multirow{2}{*}{OF-4b} & OpenAI CLIP(ViT-L/14)   & 45.6          & \textbf{93.8} & \multicolumn{1}{c|}{66.4}          & 43.3          & \textbf{84.8} & 48.8          & \textbf{50.4} & \textbf{79.2} & \multicolumn{1}{c|}{60.7}          & 25.3          & 56.5          & 35.6          \\
                                 & OVT-CLIP(ViT-L/14)      & \textbf{47.6} & \textbf{93.8} & \multicolumn{1}{c|}{\textbf{68.4}} & \textbf{44.4} & 81.5          & \textbf{51.7} & 50.1          & 76.7          & \multicolumn{1}{c|}{\textbf{62.2}} & \textbf{29.8} & \textbf{61.0} & \textbf{43.5} \\ \hline
\end{tabular}}
\vspace{-0.2cm}
\label{exp:vqa}
\end{table*}

\section{Additional Experimental results}

\begin{figure}[t]
  \centering
  \includegraphics[height=7.1cm]{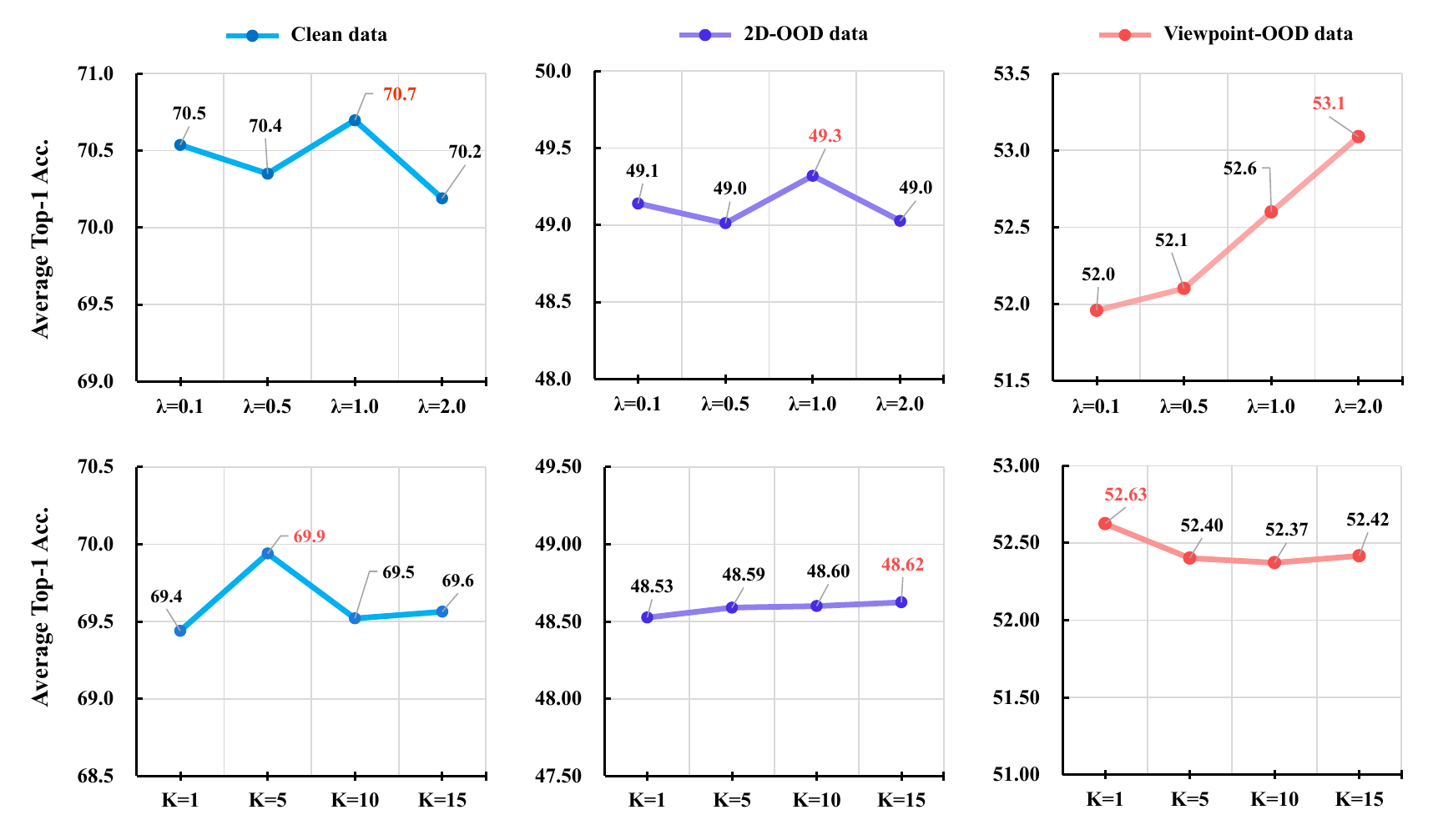}
  \caption{The curves of Top-1 average accuracy (\%) for OVT-OpenCLIP (ViT-B/32) under various data distributions, with different settings of $\lambda$ and $K$.}
  \label{fig:vis2}
  \vspace{-0.2cm}
\end{figure}

\subsection{Ablation study on $\lambda$ and $K$}
In this section, we conduct an ablation study focusing on key hyperparameters within the Omniview-Tuning (OVT) framework --- the loss balance parameter $\lambda$ and the number of outlier samples $K$ set for each object during the maximization process. We train OVT-OpenCLIP (ViT-B/32) under different ablation settings, evaluating their average Top-1 accuracy across three data distributions as in Sec.~\textcolor{red}{5.1}. For the ablation experiments on $\lambda$, we fix $K$ at 5, and for the ablation experiments on $K$, we set $\lambda$ to 1.0. All other training parameters are set consistently across each experiments, ensuring all other training parameters remain consistent across each set of experiments.

\textbf{Effects of $\lambda$:} As a balancing parameter between $\mathcal{L}_{VC}$ and $\mathcal{L}_{ITC}$, $\lambda$ critically influences the contribution ratio of these two loss terms during the fine-tuning process. Specifically, higher $\lambda$ values emphasize enhancing cross-viewpoint alignment, theoretically improving the model's performance on viewpoint shift samples. As illustrated in the first row of~\cref{fig:vis2}, where the curve on the right shows the average accuracy for viewpoint-OOD data, increasing $\lambda$ generally correlates with better performance. However, for clean and 2D-OOD samples, a higher $\lambda$ value might lead to a performance decrement. Considering the performance across three data distributions, setting $\lambda$ to 1.0 allows the model to achieve the most balanced performance. It not only realizes the highest average Top-1 accuracy on clean and 2D-OOD samples (70.7\% and 49.3\%, respectively) but also attains a 52.6\% average Top-1 accuracy on viewpoint-OOD data.

\textbf{Effects of $K$:} As shown in the second row of the curves in~\cref{fig:vis2}, the model exhibits the best performance for the clean dataset when $K$=5, reaching an average Top-1 accuracy of 69.9\%. For the 2D-OOD dataset, although there is a positive correlation between the K value and performance, the impact of the K value on performance is relatively minor, with less than 0.1\% difference in performance between $K$=15 and $K$=1. On the viewpoint OOD dataset, smaller $K$ values performed better. This can be attributed to the fact that when the number of focused outlier samples is reduced, these outliers are more likely to represent the most extreme viewpoint changes, thereby improving the model's generalization ability and consistency across different viewpoint-OOD data. Based on these experimental results, setting $ K $ to 5 is reasonable, achieving a more balanced performance across different data distributions.

\subsection{Comparison with Random Viewpoints Sampling Baselines}
Following the experimental logic of VIAT~\cite{ruan2023towards}, we compare OVT with two potential baseline methods that employ random viewpoint sampling. In OVT, random viewpoint sampling primarily considers the following two scenarios: 

\noindent\textbf{(A) Random Outlier Viewpoint Sampling (OVT-ROS).} The process of selecting outlier viewpoints is not based on a ranking of distance metrics, but rather involves randomly picking from all possible viewpoints of an object.

\noindent\textbf{(B) Random Anchor \& Outlier Viewpoint Sampling (OVT-RAOS).} Building on baseline A, further involves randomly selecting anchor viewpoints. An anchor viewpoint can be any viewpoint on the same object, not specifically the central point of viewpoint embeddings. This setting corresponds to the naive cross-viewpoint alignment method described in Sec.~\textcolor{red}{4.2}, Eq.~(\textcolor{red}{6}).

Based on the results from~\cref{exp:random}, under the condition of the same number of sampled viewpoints, the OVT method employing the min-max optimization strategy outperforms the random sampling-based OVT baseline across various data distributions. In terms of overall average Top-1 accuracy, OVT achieves a 0.7\% improvement over OVT-RAOS and a 0.6\% increase compared to OVT-ROS. Particularly in the case of viewpoint-OOD data, the average accuracy of OVT improves by 0.7\% compared to OVT-RAOS and by 1.3\% compared to OVT-ROS, demonstrating its clear advantage.

\begin{table*}[t]
\caption{Comparison between OVT and the random viewpoint sampling OVT versions (OVT-ROS and OVT-RO\&AS) within OpenCLIP (ViT-B/32). We report the average Top-1/Top-5 zero-shot accuracy (\%) under different data distributions.}

\setlength\tabcolsep{2.3pt}
\renewcommand\arraystretch{1.2}
\centering
\resizebox{\textwidth}{!}{\begin{tabular}{c|cc|cc|cc|cc}
\hline
\textbf{Method} & \multicolumn{2}{c|}{\textbf{Total Avg.}}  & \multicolumn{2}{c|}{\textbf{Clean Avg.}}  & \multicolumn{2}{c|}{\textbf{2D-OOD Avg.}} & \multicolumn{2}{c}{\textbf{Viewpoint-OOD Avg.}} \\ \hline
OVT-RAOS       & 56.8                & 82.8                & 69.7                & 91.4                & 48.8                & 76.0                & 51.9                   & 81.0                   \\
OVT-ROS         & 56.9 (\textcolor{Green}{$\uparrow$0.1})          & 82.9 (\textcolor{Green}{$\uparrow$0.1})          & 70.2 (\textcolor{Green}{$\uparrow$0.5})          & 91.7 (\textcolor{Green}{$\uparrow$0.3})          & 49.1 (\textcolor{Green}{$\uparrow$0.3})          & 76.2 (\textcolor{Green}{$\uparrow$0.2})          & 51.5 (\textcolor{Maroon}{$\downarrow$0.4})             & 80.8 (\textcolor{Maroon}{$\downarrow$0.2})             \\
OVT             & \textbf{57.5 (\textcolor{Green}{$\uparrow$0.7})} & \textbf{83.5 (\textcolor{Green}{$\uparrow$0.7})} & \textbf{70.7 (\textcolor{Green}{$\uparrow$1.0})} & \textbf{91.8 (\textcolor{Green}{$\uparrow$0.4})} & \textbf{49.3 (\textcolor{Green}{$\uparrow$0.5})} & \textbf{76.5 (\textcolor{Green}{$\uparrow$0.5})} & \textbf{52.6 (\textcolor{Green}{$\uparrow$0.7})}    & \textbf{82.3 (\textcolor{Green}{$\uparrow$1.3})}    \\ \hline
\end{tabular}
}
\label{exp:random}
\end{table*}

\begin{figure}[t]
  \centering
  \includegraphics[height=5.1cm]{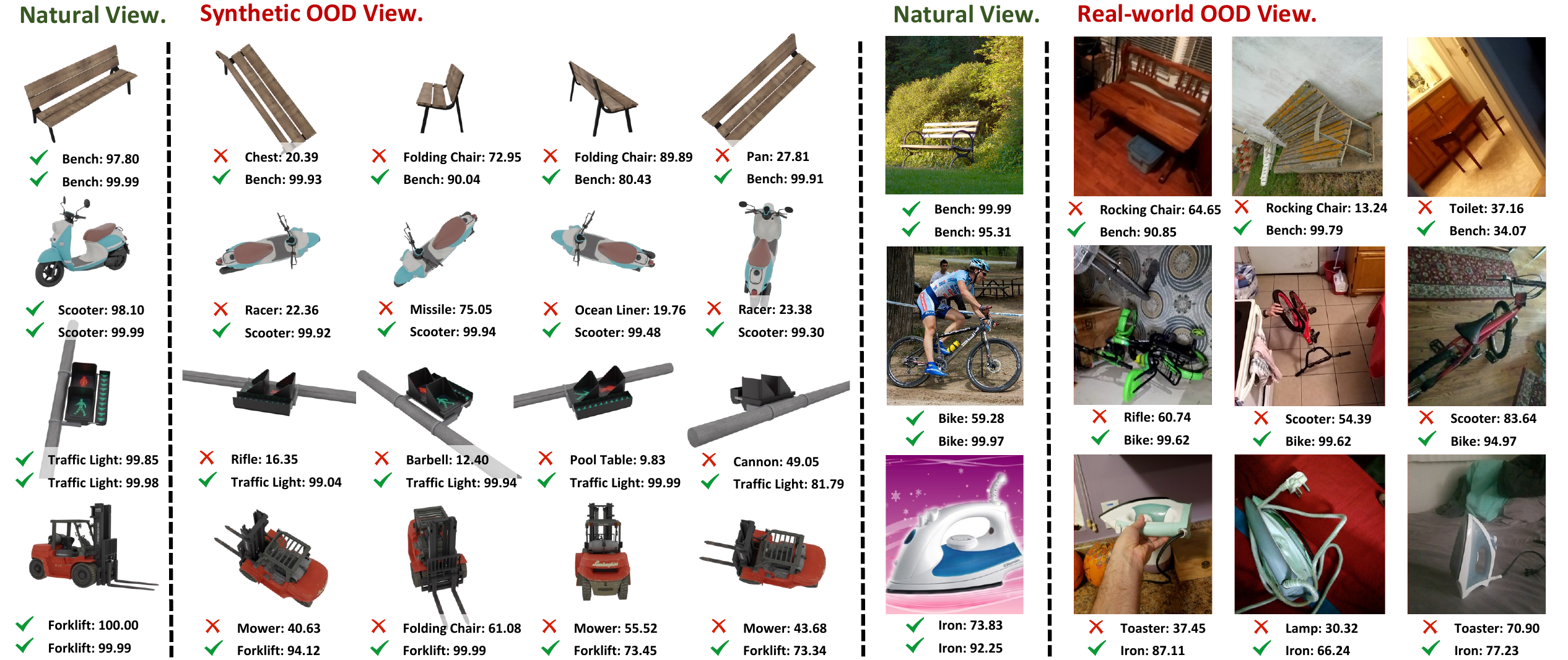}
  \caption{\textbf{Additional Visualization for zero-shot classification. } Below each image, we show the predicted categories and their confidence levels (\%) by the OpenCLIP(ViT-B/16) (\emph{first row}) and by our improved OVT-OpenCLIP(ViT-B/16) (\emph{second row}). \includegraphics[height=0.5em]{fig/correct.png} indicates a correct prediction while \includegraphics[height=0.5em]{fig/fork.png} indicating an incorrect one.
  }
  \label{fig:vis_add}
\end{figure}

\subsection{Additional visualisation results}
We provide more examples of zero-shot classification tasks, as shown in~\cref{fig:vis_add}.

\section{Explanation of the Evaluation Metrics}
In our evaluation of image captioning and VQA tasks, we designed the "Description Accuracy" metric (as seen in Tab.~\textcolor{red}{4} and~\cref{exp:vqa}). This metric calculates the semantic similarity between the category-related vocabulary contained in the captions or answers and the ground-truth category labels by utilizing a word embedding model, and it counts the proportion of samples that exceed a certain similarity threshold. To clarify this process, we formally define Description Accuracy here. Let $T^g$ be the category description text generated by the VLLMs, and $T^t$ be the ground-truth text. We use MPNet~\cite{song2020mpnet}, denoted as $\mathcal{M}$, to map these texts into the embedding space and calculate the cosine similarity between these embedding vectors. Finally, Description Accuracy is defined as the proportion of samples that meet the condition under different similarity thresholds $\beta$ as follow:
\begin{equation}
    Acc@\beta = \frac{1}{N} \cdot  \textstyle \sum_{i=1}^{N} \sigma ( \frac{\mathcal{M}(T_i^g)\cdot\mathcal{M}(T_i^t)}{\left \| \mathcal{M}(T_i^g) \right \|  \cdot \left \|\mathcal{M}(T_i^t)\right \|}  \ge \beta ),
\end{equation}
\noindent where $\sigma(\cdot)$ is an indicator function that returns 1 if the condition is true and 0 otherwise.
\begin{algorithm}[t]
\small
 \caption{\small Omniview-Tuning Algorithm}	\label{alg1}
 \KwIn{Multi-view image-text pairs $\tilde{\mathcal{D}} =\{\left \langle I_{ij}, T_{ij} \right \rangle \mid i=1,2,...,N;j=1,2,...,M_i\}$, learnable parameters $\mathbf{A},\mathbf{B},\boldsymbol{\theta}$, image encoder $E_{\mathbf{W_v}}$, text encoder $E_{\mathbf{W_t}}$, learning rate $\eta$, balance parameters $\lambda$, outlier sample size $K$.}
 \KwOut {Optimal parameters $\tilde{\mathbf{A}},\tilde{\mathbf{B}},\tilde{\boldsymbol{\theta}}$. }
Initialize $\mathbf{A},\mathbf{B},\boldsymbol{\theta}$ \;
\For{Each fine-tuning epoch}{
\tcc{Inner Maximization Step}
Calculate image embeddings $\tilde{z}^I_{ij}$ for each $I_{ij}$ by Eq.(10) \;
Calculate anchor embeddings $\tilde{z}^I_{C_i}$ for each object $i$ by Eq.(8) \;
Obtain outlier viewpoints indexes  ${\{j_1,j_2,...j_K\}} \leftarrow \max_{\{j_1,...j_K\}} d(\tilde{z}^I_{ij}, \tilde{z}^I_{C_i})$ \;
$\mathcal{O}\!=\!\{O_i\}_{i=1}^N; ~ O_i\leftarrow\{ij_1,ij_2,...,ij_K\}$ \;
\tcc {Outer minimization step}
\For{Each mini-batch}{
Calculate $\mathcal{L}_{ITC}$ by Eq.(3) \;
\eIf{$\exists~ij \in \mathcal{O}$ }{
Calculate $\mathcal{L}_{VC}$ by Eq.(7) \;
}{
$\mathcal{L}_{VC} \leftarrow 0$
}
Calculate $\mathcal{L} \leftarrow \mathcal{L}_{ITC}+\lambda\cdot\mathcal{L}_{VC}$
$\mathbf{A}\leftarrow \mathbf{A}+\eta\cdot\frac{\partial \mathcal{L} }{\partial \mathbf{A}};~\mathbf{B}\leftarrow \mathbf{B}+\eta\cdot\frac{\partial \mathcal{L} }{\partial \mathbf{B}};~\boldsymbol{\theta}\leftarrow \boldsymbol{\theta}+\eta\cdot\frac{\partial \mathcal{L} }{\partial \boldsymbol{\theta}}$ 
}
}

\end{algorithm}

\section{Pseudo-Code and Computational Cost}
To facilitate the understanding of the OVT training process, we provide the pseudocode for OVT as shown in~\cref{alg1}. In our experiments, the computational cost of the OVT fine-tuning process is primarily affected by the scale of the vision encoder and the batch size. Taking the MVCap dataset as an example, when using the ViT-B encoder, we set the batch size to 512. The outer maximization step of each fine-tuning cycle takes about 4 GPU hours, with the majority of this time occupied by the forward inference of multi-view embeddings while computing the anchor embeddings and outlier samples takes about 10 to 15 GPU minutes. The subsequent inner minimization step requires approximately 8 GPU hours. When using the ViT-L encoder and setting the batch size to 256, the maximization phase of each cycle takes about 20 GPU hours, and the minimization phase is about 40 GPU hours. The GPUs used in our experiments are the NVIDIA RTX 6000 Ada Generation with 48GB  memory.

%
%
\bibliographystyle{splncs04}
\bibliography{egbib}
\end{document}